\documentclass[10pt,twocolumn,letterpaper]{article}

\usepackage[pagenumbers]{cvpr} 

\usepackage{graphicx}
\usepackage{amsmath}
\usepackage{amssymb}
\usepackage{booktabs}
\usepackage{multirow}
\usepackage{arydshln}
\usepackage{comment}
\usepackage{enumitem}
\usepackage[table,x11names]{xcolor}

\usepackage{hyperref}
\hypersetup{pagebackref,breaklinks,colorlinks}

\usepackage[capitalize]{cleveref}
\crefname{section}{Sec.}{Secs.}
\Crefname{section}{Section}{Sections}
\Crefname{table}{Table}{Tables}
\crefname{table}{Tab.}{Tabs.}

\def\cvprPaperID{129} 
\def\confName{CVPR}
\def\confYear{2023}

\newcommand{\ut}[1]{{\color{red} #1}}

\begin{document}

\title{Towards Universal Fake Image Detectors that\\Generalize Across Generative Models}
\author{Utkarsh Ojha\thanks{Equal contribution} \hspace{20pt} Yuheng Li\footnotemark[1]  \hspace{20pt} Yong Jae Lee\\\\
 University of Wisconsin-Madison\\
}

\maketitle


\begin{abstract}
\vspace{-5pt}
With generative models proliferating at a rapid rate, there is a growing need for general purpose fake image detectors. In this work, we first show that the existing paradigm, which consists of training a deep network for real-vs-fake classification, fails to detect fake images from newer breeds of generative models when trained to detect GAN fake images. Upon analysis, we find that the resulting classifier is asymmetrically tuned to detect patterns that make an image fake. The real class becomes a `sink' class holding anything that is not fake, including generated images from models not accessible during training. Building upon this discovery, we propose to perform real-vs-fake classification \emph{without learning}; i.e., using a feature space not explicitly trained to distinguish real from fake images. We use nearest neighbor and linear probing as instantiations of this idea. When given access to the feature space of a large pretrained vision-language model, the very simple baseline of nearest neighbor classification has surprisingly good generalization ability in detecting fake images from a wide variety of generative models; e.g., it improves upon the SoTA~\cite{cnn-detect} by \textbf{+15.07 mAP} and \textbf{+25.90\% acc} when tested on unseen diffusion and autoregressive models. Our code, models, and data can be found at \url{https://github.com/Yuheng-Li/UniversalFakeDetect}
%


\end{abstract}
\vspace{-5pt}
\section{Introduction}\label{sec:intro}
\vspace{-1pt}

\begin{figure}[]
    \centering
    \includegraphics[width=0.47\textwidth]{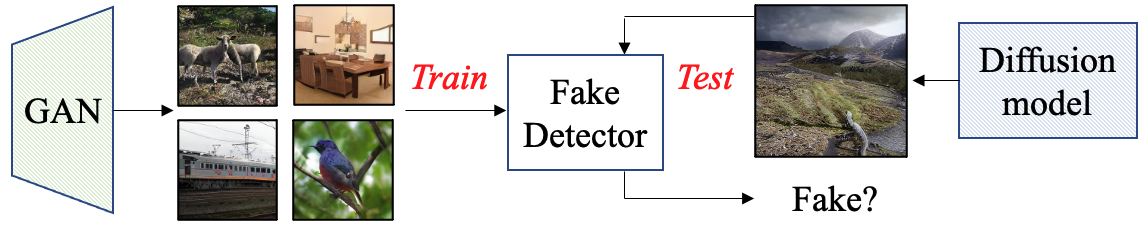}
    \caption{Using images from just one generative model, can we detect images from a different type of generative model as fake?}
    \label{fig:teaser}
    \vspace{-1em}
\end{figure}

The digital world finds itself being flooded with many kinds of fake images these days. Some could be natural images that are doctored using tools like Adobe Photoshop \cite{adobe-photoshop, detect-photoshop}, while others could have been generated through a machine learning algorithm. With the rise and maturity of deep generative models \cite{stylegan, diff-beat-gans, dall-e2}, fake images of the latter kind have caught our attention. They have raised excitement because of the quality of images one can generate with ease. They have, however, also raised concerns about their use for malicious purposes \cite{face-swap}. To make matters worse, there is no longer a single source of fake images that needs to be dealt with: for example, synthesized images could take the form of  realistic human faces generated using generative adversarial networks \cite{stylegan}, or they could take the form of complex scenes generated using diffusion models \cite{dall-e2, stable-diffusion}. One can be almost certain that there will be more modes of fake images coming in the future.  With such a diversity, our goal in this work is to develop a general purpose fake detection method which can detect whether any arbitrary image is fake, given access to only one kind of generative model during training; see Fig.~\ref{fig:teaser}.

A common paradigm has been to frame fake image detection as a learning based problem \cite{cnn-detect, patchforensics}, in which a training set of fake and real images are assumed to be available. A deep network is then trained to perform real vs fake binary classification. During test time, the model is used to detect whether a test image is real or fake. Impressively, this strategy results in an excellent generalization ability of the model to detect fake images from different algorithms \emph{within the same generative model family} \cite{cnn-detect}; e.g., a classifier trained using real/fake images from ProGAN \cite{progan} can accurately detect fake images from StyleGAN \cite{stylegan} (both being GAN variants). However, to the best of our knowledge, prior work has not thoroughly explored generalizability across different families of generative models, especially to ones not seen during training; e.g., will the GAN fake classifier be able to detect fake images from diffusion models as well? Our analysis in this work shows that existing methods \emph{do not} attain that level of generalization ability.


Specifically, we find that these models work (or fail to work) in a rather interesting manner. Whenever an image contains the (low-level) fingerprints~\cite{zhang2019gan, Yu_2019_ICCV, frank_freq, cnn-detect} particular to the generative model used for training (e.g., ProGAN), the image gets classified as fake. \emph{Anything else} gets classified as real. There are two implications: (i) even if diffusion models have a fingerprint of their own, as long as it is not very similar to GAN's fingerprint, their fake images get classified as real; (ii) the classifier doesn't seem to look for features of the real distribution when classifying an image as real; instead, the real class becomes a `sink class' which hosts anything that is not GAN's version of fake image. In other words, the decision boundary for such a classifier will be closely bound to the particular fake domain.


We argue that the reason that the classifier's decision boundary is unevenly
bound to the fake image class is because it is easy for the classifier to latch onto the low-level image artifacts that differentiate fake images from real images. Intuitively, it would be easier to learn to spot the fake pattern to classify an image as fake, rather than to learn all the ways in which an image could be real. To rectify this undesirable behavior, we propose to perform real-vs-fake image classification using features that are \emph{not trained} to separate fake from real images. As an instantiation of this idea, we perform classification using the \emph{fixed} feature space of a CLIP-ViT \cite{vit, clip} model pre-trained on internet-scale image-text pairs.  We explore both nearest neighbor classification as well as linear probing on those features.


We empirically show that our approach can achieve significantly better generalization ability in detecting fake images. For example, when training on real/fake images associated with ProGAN~\cite{progan} and evaluating on unseen diffusion and autoregressive model (LDM+Glide+Guided+DALL-E) images, we obtain improvements over the SoTA~\cite{cnn-detect} by (i) \textbf{+15.05mAP and +25.90\% acc} with nearest neighbor and (ii) \textbf{+19.49mAP and +23.39\% acc} with linear probing.
We also study the ingredients that make a feature space effective for fake image detection. For example, can we use any image encoder's feature space? Does it matter what domain of fake/real images we have access to? How large should the training feature bank be for the real/fake classes?  Our key takeaways are that while our approach is robust to the breed of generative model one uses to create the feature bank (e.g., GAN data can be used to detect diffusion models' images and vice versa), one needs the image encoder to be trained on internet-scale data (e.g., ImageNet \cite{imagenet} does not work). 


In sum, our main contributions are: (1) We analyze the limitations of existing deep learning based methods in detecting fake images from unseen breeds of generative models. (2) After empirically demonstrating prior methods' ineffectiveness, we present our theory of what could be wrong with the existing paradigm. (3) We use that analysis to present two very simple baselines for real/fake image detection: nearest neighbor and linear classification. Our approach results in state-of-the-art generalization performance, which even the oracle version of the baseline (tuning its confidence threshold on the \emph{test set}) fails to reach. (4) We thoroughly study the key ingredients of our method which are needed for good generalizability. 




\vspace{-1pt}
\section{Related work}
\vspace{-1pt}


\paragraph{Types of synthetic images.} One category involves altering a portion of a real image, and contains methods which can change a person's attribute in a source image (e.g., smile) using Adobe's photoshop tool \cite{park2020swapping, adobe-photoshop}, or methods which can create DeepFakes replacing the original face in a source image/video with a target face \cite{deepfakelab, dfaker}. Another recent technique which can optionally alter a part of a real image is DALL-E 2 \cite{dall-e2}, which can insert an object (e.g., a chair) in an existing real scene (e.g., office). The other category deals with any algorithm which generates all pixels of an image from scratch. The input for generating such images could be random noise \cite{progan, stylegan},  categorical class information \cite{brock2018biggan}, text prompts \cite{dall-e2, ldm, glide}, or could even by a collection of images \cite{li-cvpr2020}. In this work, we consider primarily this latter category of generated images and see if different detection methods can classify them as fake.


\vspace{-10pt}
\paragraph{Detecting synthetic images.} The need for detecting fake images has existed even before we had powerful image generators. When traditional methods are used to manipulate an image, the alteration in the underlying image statistics can be detected using hand-crafted cues such as compression artifacts \cite{shruti2017jpeg}, resampling \cite{alin2005resampling} or irregular reflections \cite{obrian2012reflections}. Several works have also studied GAN synthesized images in their frequency space and have demonstrated the existence of much clearer artifacts \cite{frank_freq, zhang2019gan}.

Learning based methods have been used to detect manipulated images as well \cite{splicebuster2015, rao2016detection, detect-photoshop}. Earlier methods studied whether one can even learn a classifier that can detect other images from the same generative model \cite{roessler2019faceforensicspp, marra2018gandetection, frank_freq}, and later work found that such classifiers do not generalize to detecting fakes from other models \cite{zhang2019gan, cozzolino2018}.
Hence, the idea of learning classifiers that generalize to other generative models started gaining attention \cite{natarajan, cozzolino_fake}. In that line of work,
\cite{cnn-detect} proposes a surprisingly simple and effective solution: the authors train a neural network on real/fake images from one kind of GAN, and show that it can detect images from other GAN models as well, if an appropriate training data source and data augmentations are used. \cite{patchforensics} extends this idea to detect patches (as opposed to whole images) as real/fake. \cite{proactive_fake} investigates a related, but different, task of predicting which of two test images is real and which one is modified (fake).  Our work analyses the paradigm of training neural networks for fake image detection, showing that their generalizability does not extend to unseen families of generative models.  Drawing on this finding, we show the effectiveness of a feature space \emph{not explicitly learned} for the task of fake image detection. 

\vspace{-1pt}
\section{Preliminaries}
\vspace{-1pt}

Given a test image, the task is to classify whether it was captured naturally using a camera (real image) or whether it was synthesized by a generative model (fake image). We first discuss the existing paradigm for this task \cite{cnn-detect, patchforensics}, the analysis of which leads to our proposed solution.


\vspace{-1pt}
\subsection{Problem setup}\label{sec:prob_setup}
\vspace{-1pt}

The authors in \cite{cnn-detect} train a convolutional network ($f$) for the task of binary real (0) vs fake (1) classification using images associated with one generative model. They train ProGAN \cite{progan} on 20 different object categories of LSUN \cite{yu15lsun}, and generate 18k fake images per category. In total, the real-vs-fake training dataset consists of 720k images (360k in \emph{real} class, 360k in \emph{fake} class). They choose ResNet-50 \cite{resnet} pretrained on ImageNet \cite{imagenet} as the fake classification network, and replace the fully connected layer to train the network for real vs fake classification with the binary cross entropy loss. During training, an intricate data augmentation scheme involving Gaussian blur and JPEG compression is used, which is empirically shown to be critical for generalization. Once trained, the network is used to evaluate the real and fake images from other generative models. For example, BigGAN \cite{brock2018biggan} is evaluated by testing whether its class-conditioned generated images ($F_{BigGAN}$) and corresponding real images ($R_{BigGAN}$: coming from ImageNet \cite{imagenet}) get classified correctly; i.e., whether $f(R_{BigGAN}) \approx 0$ and $f(F_{BigGAN}) \approx 1$. Similarly, each generative model (discussed in more detail in Sec.~\ref{sec:gen_models}) has a test set with an equal number of real and fake images associated with it.

\vspace{-1pt}
\subsection{Analysis of why prior work fails to generalize}\label{sec:baseline_analysis}
\vspace{-1pt}

We start by studying the ability of this network---which is trained to distinguish ProGAN fakes from real images---to detect generated images from unseen methods. In Table~\ref{tab:table_baseline}, we report the accuracy of classifying the real and fake images associated with different families of generative models. As was pointed out in \cite{cnn-detect}, when the target model belongs to the same breed of generative model used for training the real-vs-fake classifier (i.e., GANs), the network shows good overall generalizability in classifying the images; e.g., GauGAN's real/fake images can be detected with 79.25\% accuracy. However, when tested on a different family of generative models, e.g., LDM and Guided (variants of diffusion models; see Sec.~\ref{sec:gen_models}), the classification accuracy drastically drops to near \emph{chance} performance!\footnote{Corresponding precision-recall curves can be found in the appendix.}

\begin{table}[t]
 {\small
    \centering
\resizebox{1.0\linewidth}{!}{
\begin{tabular}{cccccc}
\toprule
          & CycleGAN & GauGAN & LDM & Guided & DALL-E \\
          
\midrule
Real acc. &    98.64      &     99.4   &    99.61 &     99.14   &  99.61      \\
Fake acc. &     62.91     &     59.1   &  3.05    &     4.67   &    4.9    \\
\hline
Average   &     80.77     &     79.25   &   51.33  &    51.9    &     52.26  \\
\hline
Chance performance   &     50.00     &     50.00   &   50.00  &    50.00    &     50.00  \\
\bottomrule
\end{tabular}}
}
\caption{Accuracy of a real-vs-fake classifier \cite{cnn-detect} trained on ProGAN images in detecting real and fake images from different types of generative models. LDM, Guided, and DALL-E represent the breeds of image generation algorithms not seen during training.\protect\footnotemark[1]}
\label{tab:table_baseline}
\end{table}


Now, there are two ways in which a classifier can achieve chance performance when the test set has an equal number of real and fake images: it can output (i) a random prediction for each test image, (ii) the same class prediction for all test images. From Table~\ref{tab:table_baseline}, we find that for diffusion models, the classifier works in the latter way, classifying \emph{almost all} images as real regardless of whether they are real (from LAION dataset \cite{laion}) or generated. Given this, it seems $f$ has learned an \emph{asymmetric} separation of real and fake classes, where for any image from either LDM (unseen fake) or LAION (unseen real), it has a tendency to disproportionately output one class (real) over the other (fake). 


\begin{figure}[t!]
    \centering
    \includegraphics[width=0.40\textwidth]{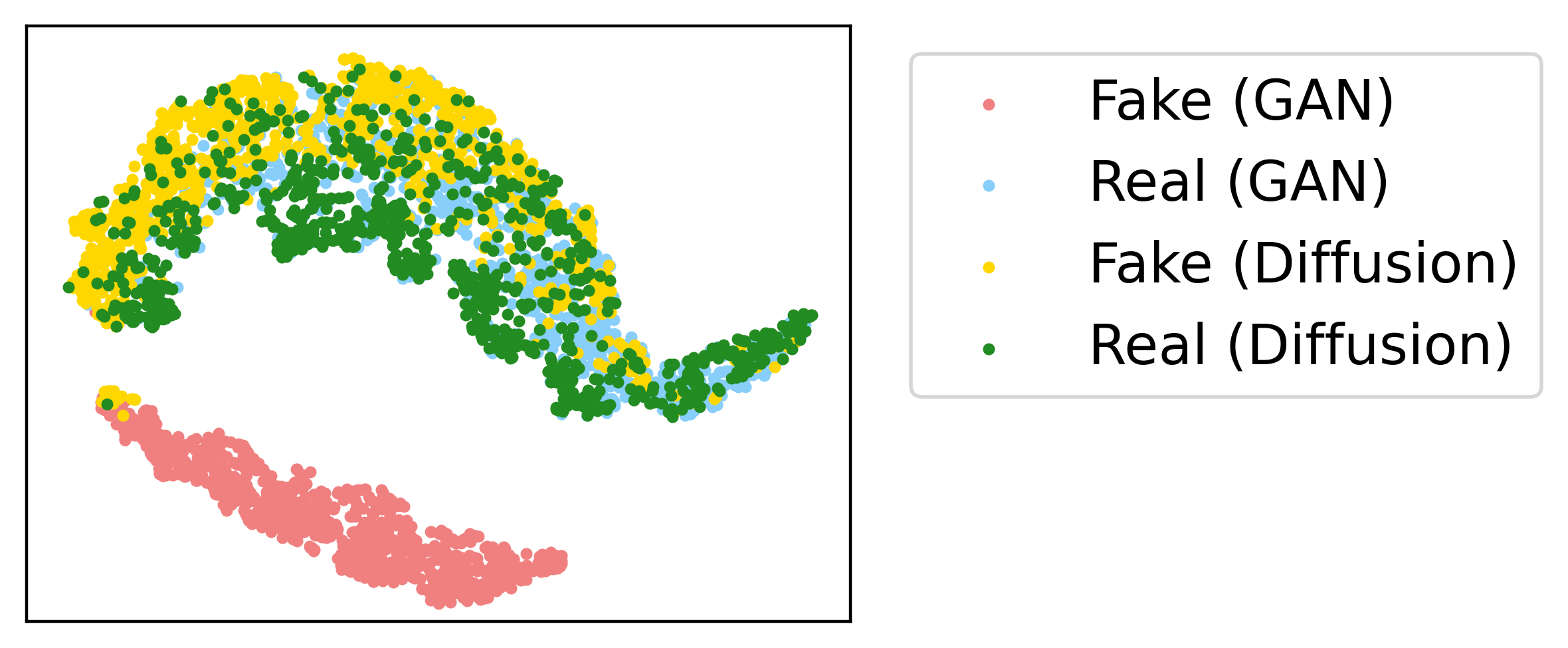}
    \caption{t-SNE visualization of real and fake images associated with two types of generative models. The feature space used is of a classifier trained to distinguish Fake (GAN) from Real (GAN).}
    \label{fig:tsne_baseline}
    \vspace{-1em}
\end{figure}

To further study this unusual phenomenon, we visualize the feature space used by $f$ for classification. We consider four image distributions: (i) $F_{GAN}$ consisting of fake images generated by ProGAN, (ii) $R_{GAN}$ consisting of the real images used to train ProGAN, (iii) $F_\mathit{Diffusion}$ consisting of fake images generated by a latent diffusion model \cite{ldm}, and (iv) $R_\mathit{Diffusion}$ consisting of real images (LAION dataset \cite{laion}) used to train the latent diffusion model. The real-vs-fake classifier is trained on (i) and (ii).  For each, we obtain their corresponding feature representations using the penultimate layer of $f$, and plot them using t-SNE \cite{tsne} in Fig.~\ref{fig:tsne_baseline}. The first thing we notice is that $f$ indeed does not treat real and fake classes equally. In the learned feature space of $f$, the four image distributions organize themselves into two noticeable clusters. The first cluster is of $F_{GAN}$ (pink) and the other is an amalgamation of the remaining three ($R_{GAN}$ + $F_\mathit{Diffusion}$ + $R_\mathit{Diffusion}$). 
In other words, $f$ can easily distinguish $F_{GAN}$ from the other three, but the learned real class does not seem to have any property (a space) of its own, but is rather used by $f$ to form a \emph{sink class}, which hosts anything that is not $F_{GAN}$.
The second thing we notice is that the cluster surrounding the learned fake class is very condensed compared to the one surrounding the learned real class, which is much more open. This indicates that $f$ can detect a common property among images from $F_{GAN}$ with more ease than detecting a common property among images from $R_{GAN}$.

\begin{figure}[t!]
    \centering
    \includegraphics[width=0.48\textwidth]{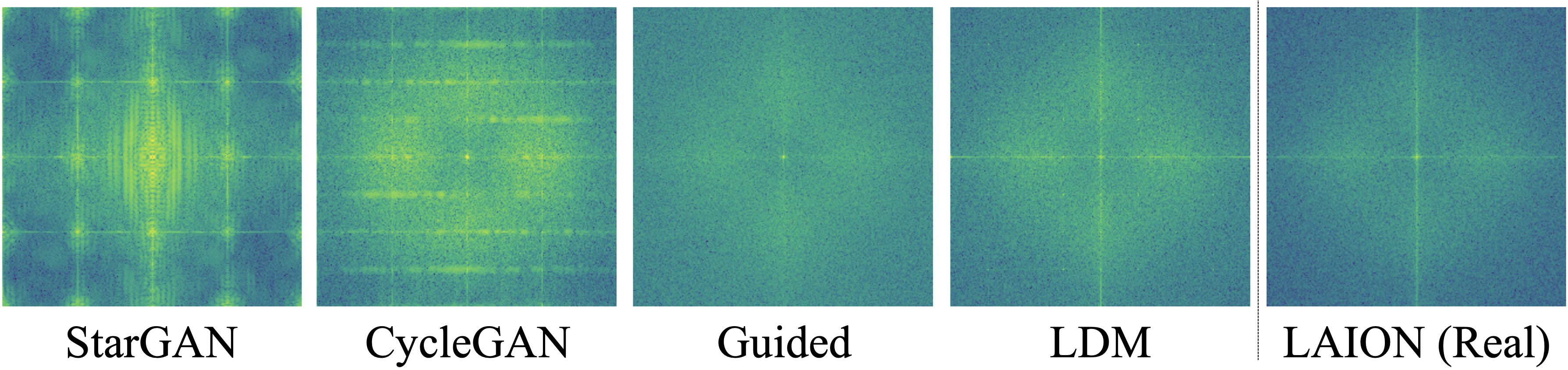}
    \caption{Average frequency spectra of each domain. The first four correspond to fake images from GANs and diffusion models. The last one represents real images from LAION~\cite{laion} dataset.}
    \label{fig:fft}
    \vspace{-1em}
\end{figure}

But why is it that the property that $f$ finds to be common among $F_{GAN}$ is useful for detecting fake images from other GAN models (e.g., CycleGAN), but not for detecting $F_\mathit{Diffusion}$? In what way are fake images from diffusion models different than images from GANs?  We investigate this by visualizing the frequency spectra of different image distributions, inspired by \cite{zhang2019gan,cnn-detect,cai-wacv2023, cai2021frequency}. For each distribution (e.g., $F_{BigGAN}$), we start by performing a high pass filtering for each image by subtracting from it its median blurred image. We then take the average of the resulting high frequency component across 2000 images, and compute the Fourier transform. Fig.~\ref{fig:fft} shows this average frequency spectra for four fake domains and one real domain. Similar to \cite{cnn-detect}, we see a distinct and repeated pattern in StarGAN and CycleGAN. However, this pattern is missing in the fake images from diffusion models (Guided \cite{guided-diffusion} and LDM \cite{ldm}), similar to images from a real distribution (LAION \cite{laion}). So, while fake images from diffusion models seem to have some common property of their own, Fig.~\ref{fig:fft} indicates that that property is not of a similar nature as the ones shared by GANs.

Our hypothesis is that when $f$ is learning to distinguish between $F_{GAN}$ and $R_{GAN}$, it latches onto the artifacts depicted in Fig.~\ref{fig:fft}, learning only to look for the presence/absence of those patterns in an image. Since this is sufficient for it to reduce the training error, it largely ignores learning any features (e.g., smooth edges) pertaining to the \emph{real} class. This, in turn, results in a skewed decision boundary where a fake image from a diffusion model, lacking the GAN's fingerprints, ends up being classified as real.



\begin{figure*}[t]
    \centering
    \includegraphics[width=.9\textwidth]{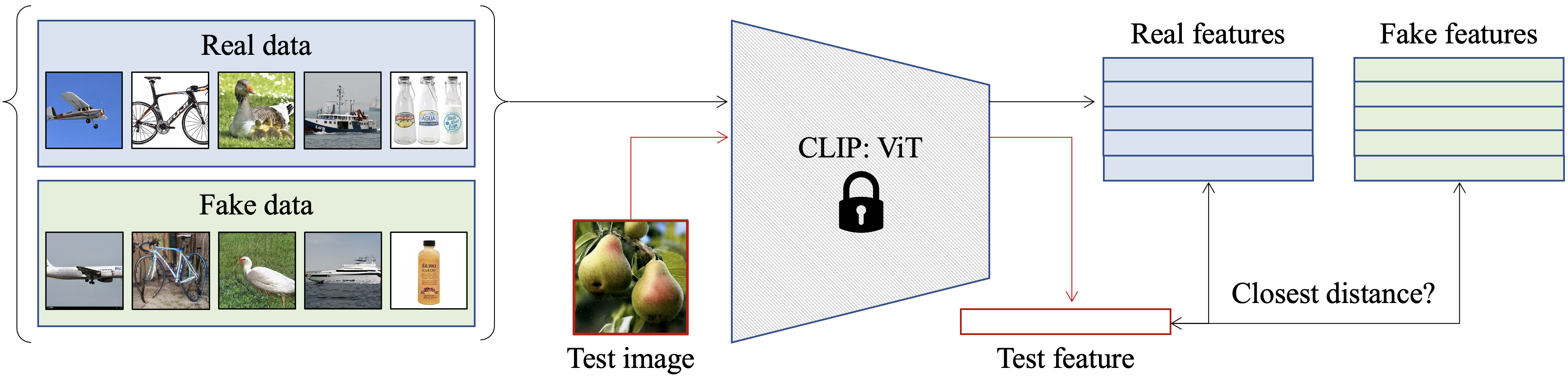}
    \caption{\textbf{Nearest neighbors for real-vs-fake classification.} We first map the real and fake images to their corresponding feature representations using a pre-trained CLIP:ViT network \emph{not trained for this task}. A test image is mapped into the same feature space, and cosine distance is used to find the closest member in the feature bank. The label of that member is the predicted class.}
    \label{fig:approach}
\end{figure*}

\vspace{-1pt}
\section{Approach}
\vspace{-1pt}

If learning a neural network $f$ is not an ideal way to separate real ($\mathcal{R}$) and fake ($\mathcal{F}$) classes, what should we do? The key, we believe, is that the classification process should happen in a feature space which has \emph{not been learned} to separate images from the two classes. This might ensure that the features are not biased to recognize patterns from one class disproportionately better than the other. 


\vspace{-10pt}
\paragraph{Choice of feature space.}
As an initial idea, since we might not want to learn any features, can we simply perform the classification in pixel space? This would not work, as pixel space would not capture any meaningful information (e.g., edges) beyond point-to-point pixel correspondences. So, any classification decision of an image should be made after it has been mapped into some feature space. This feature space, produced by a network and denoted as $\phi$, should have some desirable qualities.

First, $\phi$ should have been exposed to a large number of images. Since we hope to design a general purpose fake image detector, its functioning should be consistent for a wide variety of real/fake images (e.g., a human face, an outdoor scene). This calls for the feature space of $\phi$ to be heavily populated with different kinds of images, so that for any new test image, it knows how to embed it properly. Second, it would be beneficial if $\phi$, while being general overall, can also capture low-level details of an image. This is because differences between real and fake images arise particularly at low-level details~\cite{patchforensics, zhang2019gan}.

To satisfy these requirements, we consider leveraging a large network trained on huge amounts of data, as a possible candidate to produce $\phi$. In particular, we choose a variant of the vision transformer, ViT-L/14 \cite{vit}, trained for the task of image-language alignment, CLIP \cite{clip}. CLIP:ViT is trained on an extraordinarily large dataset of 400M image-text pairs, so it satisfies the first requirement of sufficient exposure to the visual world. Additionally, since ViT-L/14 has a smaller starting patch size of 14 $\times$ 14 (compared to other ViT variants), we believe it can also aid in modeling the low-level image details needed for real-vs-fake classification. Hence, for all of our main experiments, we use the last layer of CLIP:ViT-L/14's visual encoder as $\phi$. 

The overall approach can be formalized in the following way. We assume access to images associated with a single generative model (e.g., ProGAN, which is the same constraint as in \cite{cnn-detect}). $\mathcal{R} = \{r_1, r_2, ..., r_N\}$, and $\mathcal{F} = \{f_1, f_2, ..., f_N\}$ denote the real and fake classes respectively, each containing $N$ images. $\mathcal{D} = \{\mathcal{R} \cup \mathcal{F}\}$ denotes the overall training set. We investigate two simple classification methods: nearest neighbor and linear probing.  Importantly, both methods utilize a feature space that is entirely untrained for real/fake classification.

\vspace{-10pt}
\paragraph{Nearest neighbor.}
Given the pre-trained CLIP:ViT visual encoder, we use its final layer $\phi$ to map the entire training data to their feature representations (of 768 dimensions). The resulting feature bank is $\phi_{bank}$ = \{$\phi_{\mathcal{R}} \cup \phi_{\mathcal{F}}$\} where $\phi_{\mathcal{R}}$ = $\{\phi_{r_1}, \phi_{r_2}, ..., \phi_{r_N}$\} and $\phi_{\mathcal{F}}$ = $\{\phi_{f_1}, \phi_{f_2}, ..., \phi_{f_N}\}$. During test time, an image $x$ is first mapped to its feature representation $\phi_{x}$. Using cosine distance as the metric $d$, we find its nearest neighbor to both the real ($\phi_{\mathcal{R}}$) and fake ($\phi_{\mathcal{F}}$) feature banks. The prediction---real:0, fake:1---is given based on the smaller distance of the two:
\begin{equation*}
  \text{pred}(x)=\begin{cases}
    1, & \text{if $\min_i{(d(\phi_{x}, \phi_{f_i}))}$ $<$ $\min_i{(d(\phi_{x}, \phi_{r_i}))}$}\\
    0, & \text{otherwise}.
  \end{cases}
\end{equation*}

The CLIP:ViT encoder is always kept frozen; see Fig.~\ref{fig:approach}. 

\vspace{-10pt}
\paragraph{Linear classification.}
We take the pre-trained CLIP:ViT encoder, and add a single linear layer with sigmoid activation on top of it, and train \emph{only} this new classification layer $\psi$ for binary real-vs-fake classification using binary cross entropy loss:
\begin{equation*}
  \mathcal{L} = -\sum_{f_i \in \mathcal{F}}\log(\psi(\phi_{f_i})) -\sum_{r_i \in \mathcal{R}}\log(1 - \psi(\phi_{r_i})).
\end{equation*}


Since such a classifier involves training only a few hundred parameters in the linear layer (e.g., 768), conceptually, it will be quite similar to nearest neighbor and retain many of its useful properties. Additionally, it has the benefit of being more computation and memory friendly.

\vspace{-1pt}
\section{Experiments}
\vspace{-1pt}

We now discuss the experimental setup for evaluating the proposed method for the task of fake image detection. 

\vspace{-1pt}
\subsection{Generative models studied}\label{sec:gen_models}
\vspace{-1pt}

Since new methods of creating fake images are always coming up, the standard practice is to limit access to only one generative model during training, and test the resulting model on images from unseen generative models. We follow the same protocol as described in \cite{cnn-detect} and use ProGAN's real/fake images as the training dataset.

During evaluation, we consider a variety of generative models. First, we evaluate on the models used in \cite{cnn-detect}: ProGAN~\cite{progan}, StyleGAN~\cite{stylegan}, BigGAN~\cite{brock2018biggan}, CycleGAN~\cite{Zhu-2017-cycleGAN}, StarGAN~\cite{choi2018stargan}, GauGAN~\cite{park2019SPADE}, CRN~\cite{crn}, IMLE~\cite{imle}, SAN~\cite{san}, SITD~\cite{sitd}, and DeepFakes~\cite{roessler2019faceforensicspp}. Each generative model has a collection of real and fake images. Additionally, we evaluate on guided diffusion model \cite{guided-diffusion}, which is trained for the task for class conditional image synthesis on the ImageNet dataset \cite{imagenet}. We also perform evaluation on recent text-to-image generation models: (i) Latent diffusion model (LDM) \cite{ldm} and (ii) Glide \cite{glide} are variants of diffusion models, and (iii) DALL-E \cite{dalle-orig} is an autoregressive model (we consider its open sourced implementation DALL-E-mini \cite{dalle-mini}). For these three methods, we set the LAION dataset \cite{laion} as the real class, and use the corresponding text descriptions to generate the fake images.

LDMs, being diffusion models, can be used to generate images in different ways. The standard practice is to use a text-prompt as input, and perform 200 steps of noise refinement. One can also generate an image with the help of guidance, or use fewer steps for faster sampling. So, we consider three variants of a pre-trained LDM for evaluation purposes: (i) LDM with 200 steps, (ii) LDM with 200 steps with classifier-free diffusion guidance (CFG), and (iii) LDM with 100 steps. Similarly, we also experiment with different variants of a pre-trained Glide model, which consists of two separate stages of noise refinement. The standard practice is to use 100 steps to get a low resolution image at 64 $\times$ 64, then use 27 steps to upsample the image to 256 $\times$ 256 in the next stage. We consider three Glide variants based on the number of refinement steps in the two stages: (i) 100 steps in the first stage followed by 27 steps in the second stage (100-27), (ii) 50-27, and (iii) 100-10. All generative models synthesize 256 $\times$ 256 resolution images.

\vspace{-1pt}
\subsection{Real-vs-Fake classification baselines}\label{sec:baselines}
\vspace{-1pt}

We compare with the following state-of-the-art baselines: (i) Training a classification network to give a real/fake decision for an image using binary cross-entropy loss \cite{cnn-detect}. The authors take a ResNet-50 \cite{resnet} pre-trained on ImageNet, and finetune it on ProGAN's real/fake images (henceforth referred as trained deep network). (ii) We include another variant where we change the backbone to CLIP:ViT \cite{vit} (to match our approach) and train the network for the same task. (iii) Training a similar classification network on a patch level instead \cite{patchforensics}, where the authors propose to truncate either a ResNet~\cite{resnet} or Xception~\cite{xception} so that a smaller receptive field is considered when making the decision. This method was primarily proposed for detecting generated \emph{facial} images, but we study whether the idea can be extended to detect more complex fake images. We consider two variants within this baseline; ResNet50-Layer1 and Xception-Block2, where Layer1 and Block2 denote the layers after which truncation is applied. (iv) Training a classification network where input images are first converted into their corresponding co-occurrence matrices \cite{natarajan} (a technique shown to be effective in image steganalysis and forensics \cite{co_occur_steg, co_occur_forensics}), conditioned on which the network predicts the real/fake class. (v) Training a classification network on the frequency spectrum of real/fake images \cite{zhang2019gan}, a space which the authors show as better in capturing and displaying the artifacts present in the GAN generated images.

All details regarding the training process of the baselines (e.g. number of training iterations, learning rates) can be found in the appendix.


\begin{table*}[t!]
  {\small
    \centering
    \tabcolsep=0.1cm
    \resizebox{1.\linewidth}{!}{
    \begin{tabular}{cc cccccc c cc cc c ccc ccc c c}
    \toprule

        \multirow{2}{*}{\shortstack[c]{Detection\\method}} & \multirow{2}{*}{Variant} & \multicolumn{6}{c}{Generative Adversarial Networks} & \multirow{2}{*}{\shortstack[c]{Deep\\fakes}} & \multicolumn{2}{c}{Low level vision} & \multicolumn{2}{c}{Perceptual loss} &\multirow{2}{*}{Guided} & \multicolumn{3}{c}{LDM} & \multicolumn{3}{c}{Glide} & \multirow{2}{*}{DALL-E} & Total \\
    \cmidrule(lr){3-8} \cmidrule(lr){10-11} \cmidrule(lr){12-13} \cmidrule(lr){15-17} \cmidrule(lr){18-20} \cmidrule(lr){22-22}

    & & \shortstack[c]{Pro-\\GAN} & \shortstack[c]{Cycle-\\GAN} & \shortstack[c]{Big-\\GAN} & \shortstack[c]{Style-\\GAN} & \shortstack[c]{Gau-\\GAN} &  \shortstack[c]{Star-\\GAN}   &  & SITD & SAN & CRN & IMLE & & \shortstack[c]{200\\steps} & \shortstack[c]{200\\w/ CFG} & \shortstack[c]{100\\steps} & \shortstack[c]{100\\27} & \shortstack[c]{50\\27} & \shortstack[c]{100\\10} & & \shortstack[c]{mAP}
    
    \\ 

    \midrule

\multirow{3}{*}{\shortstack[c]{Trained\\deep network \cite{cnn-detect}}} & Blur+JPEG (0.1) & \textbf{100.0} & 93.47 & 84.5 & \textbf{99.54} & 89.49 & 98.15 & 89.02 & 73.75 & 59.47 & 98.24 & 98.4 & 73.72 & 70.62 & 71.0 & 70.54 & 80.65 & 84.91 & 82.07 & 70.59 & 83.58 \\
& Blur+JPEG (0.5) & \textbf{100.0} & 96.83 & 88.24 & 98.29 & 98.09 & 95.44 & 66.27 & 86.0 & 61.2 & \textbf{98.94} & \textbf{99.52} & 68.57 & 66.0 & 66.68 & 65.39 & 73.29 & 78.02 & 76.23 & 65.93 & 81.52\\
& ViT:CLIP (B+J 0.5) &  99.98 & 93.32 & 83.63 & 88.14 & 92.81 & 84.62 & 67.23 & \textbf{93.48} & 55.21 & 88.75 & 96.22 & 55.74 & 52.52 & 54.51 & 52.2 & 56.64 & 61.13 & 56.64 & 62.74 & 73.44 \\ 

\midrule

\multirow{2}{*}{\shortstack[c]{Patch\\classifier \cite{patchforensics}}} & ResNet50-Layer1  &  98.86 & 72.04 & 68.79 & 92.96 & 55.9 & 92.06 & 60.18 & 65.82 & 52.87 & 68.74 & 67.59 & 70.05 & 87.84 & 84.94 & 88.1 & 74.54 & 76.28 & 75.84 & 77.07 & 75.28\\
& Xception-Block2 &  80.88 & 72.84 & 71.66 & 85.75 & 65.99 & 69.25 & 76.55 & 76.19 & 76.34 & 74.52 & 68.52 & 75.03 & 87.1 & 86.72 & 86.4 & 85.37 & 83.73 & 78.38 & 75.67 & 77.73\\

\midrule

\multirow{1}{*}{\shortstack[c]{Co-occurence~\cite{natarajan}}} & -  &  99.74 & 80.95 & 50.61 & 98.63 & 53.11 & 67.99 & 59.14& 68.98 & 60.42 & 73.06 & 87.21 & 70.20 & 91.21 & 89.02 & 92.39 & 89.32 & 88.35 & 82.79 & 80.96 & 78.11\\

\midrule

\shortstack[c]{Freq-spec \cite{zhang2019gan}} & CycleGAN & 55.39 & \textbf{100.0} & 75.08 & 55.11 & 66.08 & \textbf{100.0} & 45.18 & 47.46 & 57.12 & 53.61 & 50.98 & 57.72 & 77.72 & 77.25 & 76.47 & 68.58 & 64.58 & 61.92 & 67.77 & 66.21\\

\midrule
\rowcolor{lightgray!30}
 & NN, $k=1$ & \textbf{100.0} & 98.14 & 94.49 & 86.68 & 99.26 & 99.53 & \textbf{93.09} & 78.46 & 67.54 & 83.13 & 91.07 & 79.31 & 95.84 & 79.84 & 95.97 & 93.98 & 95.17 & \textbf{96.05} & 88.51 & 90.32\\
\rowcolor{lightgray!30}
& NN, $k=3$ & \textbf{100.0} & 98.13 & 94.46 & 86.67 & 99.25 & 99.53 & 93.03 & 78.54 & 67.54 & 83.13 & 91.06 & 79.26 & 95.81 & 79.78 & 95.94 & 93.94 & 95.13 & 94.60 & 88.47 & 90.22 \\
\rowcolor{lightgray!30}
& NN, $k=5$ & \textbf{100.0} & 98.13 & 94.46 & 86.66 & 99.25 & 99.53 & 93.02 & 78.54 & 67.54 & 83.12 & 91.06 & 79.25 & 95.81 & 79.78 & 95.94 & 93.94 & 95.13 & 94.60 & 88.46 & 90.22 \\
\rowcolor{lightgray!30}
& NN, $k=9$ & \textbf{100.0} & 98.13 & 94.46 & 86.66 & 99.25 & 99.53 & 91.67 & 78.54 & 67.54 & 83.12 & 91.06 & 79.24 & 95.81 & 79.77 & 95.93 & 93.93 & 95.12 & 94.59 & 88.45 & 90.14\\
\rowcolor{lightgray!30}
\multirow{-5}{*}{\shortstack[c]{\textbf{Ours}}} & LC & \textbf{100.0} & 99.46 & \textbf{99.59} & 97.24 & \textbf{99.98} & 99.60 & 82.45 & 61.32 & \textbf{79.02} & 96.72 & 99.00 & \textbf{87.77} & \textbf{99.14} & \textbf{92.15} & \textbf{99.17} & \textbf{94.74} & \textbf{95.34} & 94.57 & \textbf{97.15} & \textbf{93.38}\\

    \bottomrule
    \end{tabular}}
    }
    \vspace{-1.5mm}
    \caption{\textbf{Generalization results.} Average precision (AP) of different methods for detecting real/fake images. Models outside the GANs column can be considered as the generalizing domain, since ProGAN data is being used as the train set. The improvements using the fixed CLIP:ViT feature backbone (Ours NN/LC) over the best performing baseline i.e., the trained deep network~\cite{cnn-detect}, is particularly noticeable when evaluating on unseen generative models (e.g., LDM), where our best performing method has significant gains over the best performing baseline: +9.8 mAP across all settings and +19.49 mAP across unseen diffusion \& autoregressive models (LDM+Glide+Guided+DALL-E).}
    \label{tab:ap}

\end{table*}

\begin{table*}[t!]
  {\small
    \centering
    \tabcolsep=0.1cm
    \resizebox{1.\linewidth}{!}{
    \begin{tabular}{cc cccccc c cc cc c ccc ccc c c}
    \toprule

        \multirow{2}{*}{\shortstack[c]{Detection\\method}} & \multirow{2}{*}{Variant} & \multicolumn{6}{c}{Generative Adversarial Networks} &\multirow{2}{*}{\shortstack[c]{Deep\\fakes}} & \multicolumn{2}{c}{Low level vision} & \multicolumn{2}{c}{Perceptual loss} &\multirow{2}{*}{Guided} & \multicolumn{3}{c}{LDM} & \multicolumn{3}{c}{Glide} & \multirow{2}{*}{DALL-E} & Total \\
    \cmidrule(lr){3-8} \cmidrule(lr){10-11} \cmidrule(lr){12-13} \cmidrule(lr){15-17} \cmidrule(lr){18-20} \cmidrule(lr){22-22}

    & & \shortstack[c]{Pro-\\GAN} & \shortstack[c]{Cycle-\\GAN} & \shortstack[c]{Big-\\GAN} & \shortstack[c]{Style-\\GAN} & \shortstack[c]{Gau-\\GAN} &  \shortstack[c]{Star-\\GAN}   &  & SITD & SAN & CRN & IMLE & & \shortstack[c]{200\\steps} & \shortstack[c]{200\\w/ CFG} & \shortstack[c]{100\\steps} & \shortstack[c]{100\\27} & \shortstack[c]{50\\27} & \shortstack[c]{100\\10} & & \shortstack[c]{Avg.\\acc}
    
    \\ 

    \midrule

\multirow{4}{*}{\shortstack[c]{Trained\\deep network \cite{cnn-detect}}} & Blur+JPEG (0.1) & 99.99 & 85.20 & 70.20 & 85.7 & 78.95 & 91.7 & 53.47 & 66.67 & 48.69 & 86.31 & 86.26 & 60.07 & 54.03 & 54.96 & 54.14 & 60.78 & 63.8 & 65.66 & 55.58 & 69.58 \\
& Blur+JPEG (0.5) & \textbf{100.0} & 80.77 & 58.98 & 69.24 & 79.25 & 80.94 & 51.06 & 56.94 & 47.73 & 87.58 & 94.07 & 51.90 & 51.33 & 51.93 & 51.28 & 54.43 & 55.97 & 54.36 & 52.26 & 64.73 \\
& Oracle$^{*}$ (B+J 0.5) & \textbf{100.0} & 90.88 & 82.40 & \textbf{93.11} & 93.52 & 87.27 & 62.48 & \textbf{76.67} & 57.04 & \textbf{95.28} & \textbf{96.93} & 65.20 & 63.15 & 62.39 & 61.50 & 65.36 & 69.52 & 66.18 & 60.10 & 76.26 \\ 
& ViT:CLIP (B+J 0.5) & 98.94 & 78.80 & 60.62 & 60.56 & 66.82 & 62.31 & 52.28 & 65.28 & 47.97 & 64.09 & 79.54 & 50.66 & 50.74 & 51.04 & 50.76 & 52.15 & 53.07 & 52.06 & 53.18 & 60.57 \\ 

\midrule

\multirow{2}{*}{\shortstack[c]{Patch\\classifier \cite{patchforensics}}} & ResNet50-Layer1  &  94.38 & 67.38 & 64.62 & 82.26 & 57.19 & 80.29 & 55.32 & 64.59 & 51.24 & 54.29 & 55.11 & 65.14 & 79.09 & \textbf{76.17} & 79.36 & 67.06 & 68.55 & 68.04 & 69.44 & 68.39\\
& Xception-Block2 &  75.03 & 68.97 & 68.47 & 79.16 & 64.23 & 63.94 & 75.54 & 75.14 & \textbf{75.28} & 72.33 & 55.3 & 67.41 & 76.5 & 76.1 & 75.77 & 74.81 & 73.28 & 68.52 & 67.91 & 71.24\\



\midrule

\multirow{1}{*}{\shortstack[c]{Co-occurence~\cite{natarajan}}} & -  & 97.70 & 63.15 & 53.75 & 92.50 & 51.1 & 54.7 & 57.1 & 63.06 & 55.85 & 65.65 & 65.80 & 60.50 & 70.7 & 70.55 & 71.00 & 70.25 & 69.60 & 69.90 & 67.55 & 66.86\\

\midrule

\shortstack[c]{Freq-spec \cite{zhang2019gan}} & CycleGAN & 49.90 & \textbf{99.90} & 50.50 & 49.90 & 50.30 & \textbf{99.70} & 50.10 & 50.00 & 48.00 & 50.60 & 50.10 & 50.90 & 50.40 & 50.40 & 50.30 & 51.70 & 51.40 & 50.40 & 50.00 & 55.45\\

\midrule
\rowcolor{lightgray!30}
 & NN, $k=1$ & 99.58 & 94.70 & 86.95 & 80.24 & 96.67 & 98.84 & 80.9 & 71.0 & 56.0 & 66.3 & 76.5 & 68.76 & 89.56 & 68.99 & 89.51 & 86.44 & 88.02 & 87.27 & 77.52 & 82.30 \\
\rowcolor{lightgray!30}
& NN, $k=3$ & 99.58 & 95.04 & 87.63 & 80.55 & 96.94 & 98.77 & 83.05 & 71.5 & 59.5 & 66.69 & 76.87 & 70.02 & 90.37 & 70.17 & 90.57 & 87.84 & 89.34 & 88.78 & 79.29 & 83.28 \\
\rowcolor{lightgray!30}
& NN, $k=5$ & 99.60 & 94.32 & 88.23 & 80.60 & 97.00 & 98.90 & 83.85 & 71.5 & 60.0 & 67.04 & 78.02 & 70.55 & 90.89 & 70.97 & 91.01 & 88.42 & 90.07 & 89.60 & 80.19 & 83.72  \\
\rowcolor{lightgray!30}
& NN, $k=9$ & 99.54 & 93.49 & 88.63 & 80.75 & 97.11 & 98.97 & \textbf{84.5} & 71.5 & 61.0 & 69.27 & 79.21 & \textbf{71.06} & 91.29 & 72.02 & 91.29 & \textbf{89.05} & \textbf{90.67} & \textbf{90.08} & 81.47 & \textbf{84.25} \\
\rowcolor{lightgray!30}
\multirow{-5}{*}{\shortstack[c]{\textbf{Ours}}} & LC & \textbf{100.0} & 98.50 & \textbf{94.50} & 82.00 & \textbf{99.50} & 97.00 & 66.60 & 63.00 & 57.50 & 59.5 & 72.00 & 70.03 & \textbf{94.19} & 73.76 & \textbf{94.36} & 79.07 & 79.85 & 78.14 & \textbf{86.78} & 81.38 \\

    \bottomrule
    \end{tabular}}
    }
    \vspace{-1.5mm}
    \caption{\textbf{Generalization results.}   Analogous result of Table~\ref{tab:ap}, where we use classification accuracy to compare the methods. Numbers indicate the average accuracy over real and fake images from a test model. Oracle with $^{*}$ indicates that the method uses the test set to calibrate the confidence threshold. Similar to Table~\ref{tab:ap}, the generalization ability of the fixed feature backbone (Ours NN/LC) can be seen in the significant gain in accuracy (+25-30\% over the baselines) when testing on unseen generative model families.}
    \label{tab:main_comparison}

\end{table*}

\vspace{-1pt}
\subsection{Evaluation metrics}\label{sec:eval_metrics}
\vspace{-1pt}

We follow existing works \cite{zhang2019gan, frank_freq, natarajan, cnn-detect, patchforensics} and report both average precision (AP) and classification accuracy. To compute classification accuracy for the baselines, we tune the classification threshold on the held-out training validation set of the available generative model.  For example, when training a classifier on data associated with ProGAN, the threshold is chosen so that the accuracy on a held out set of ProGAN's real and fake images can be maximized. In addition, we also compute an upper-bound \emph{oracle} accuracy for \cite{cnn-detect}, where the classifier's threshold is calibrated directly on each test set separately. This is to gauge the best that the classifier could have performed on each test set. The details of tuning the threshold are explained in the appendix.

\begin{figure*}[t]
    \centering
    \includegraphics[width=.98\textwidth]{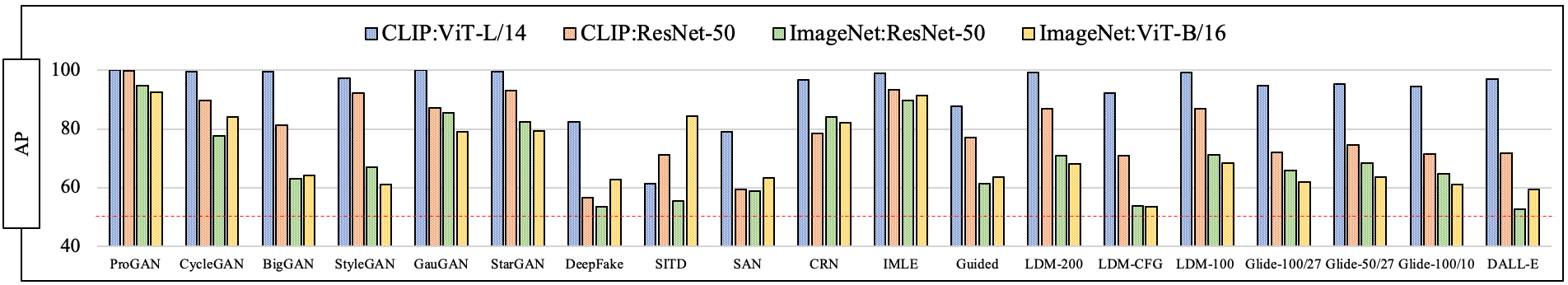}
    \caption{\textbf{Ablation on the network architecture and pre-training dataset.} A network trained on the task of CLIP is better equipped at separating fake images from real, compared to networks trained on ImageNet classification. The red dotted line depicts chance performance.}
    \label{fig:pretrain_ablation}
\end{figure*}

\vspace{-1pt}
\section{Results}
\vspace{-1pt}

We start by comparing our approach to the state-of-the-art baselines in their ability to classify real/fake images from a suite of generative models. We then study the different components of our approach, e.g., the effect of network architecture, size of the feature bank for nearest neighbor.

\subsection{Detecting fake images from unseen methods}

Table~\ref{tab:ap} and Table~\ref{tab:main_comparison} show the average precision (AP) and classification accuracy, respectively, of all methods (rows) in detecting fake images from different generative models (columns). For classification accuracy, the numbers shown are averaged over the real and fake classes for each generative model.\footnote{\label{note1}See appendix which further breaks down the accuracies for real/fake.} 
All methods have access to only ProGAN's data (except \cite{zhang2019gan}, which uses CycleGAN's data), either for training the classifier or for creating the nearest neighbor feature bank. 

As discussed in Sec.~\ref{sec:baseline_analysis}, the trained classifier baseline~\cite{cnn-detect} distinguishes real from fakes with good accuracy for other GAN variants. However, the accuracy drops drastically (sometimes to nearly chance performance $\sim$50-55\%; e.g., LDM variants) for images from most unseen generative models, where all types of fake images are classified mostly as real (please see Table C in the supplementary). Importantly, this behavior does not change even if we change the backbone to CLIP:ViT (the one used by our methods). This tells us that the issue highlighted in Fig.~\ref{fig:tsne_baseline} affects deep neural networks in general, and not just ResNets.  In fact, CLIP:ViT performs slightly worse than using a ResNet, which shows that the higher the capacity, the easier it is for that model to overfit to the fake artifacts during training. Performing classification on a patch-level \cite{patchforensics}, using co-occurence matrices \cite{natarajan}, or using the frequency space \cite{zhang2019gan} does not solve the issue either, where the classifier fails to have a consistent detection ability, sometimes even for methods within the same generative model family (e.g., GauGAN/BigGAN). Furthermore, even detecting real/fake patches in images from the same training domain (ProGAN) can be difficult in certain settings (Xception). This indicates that while learning to find patterns within small image regions might be sufficient when patches do not vary too much (e.g., facial images), it might not be sufficient when the domain of real and fake images becomes more complex (e.g., natural scenes). 


\begin{figure}[t]
    \centering
    \includegraphics[width=0.48\textwidth]{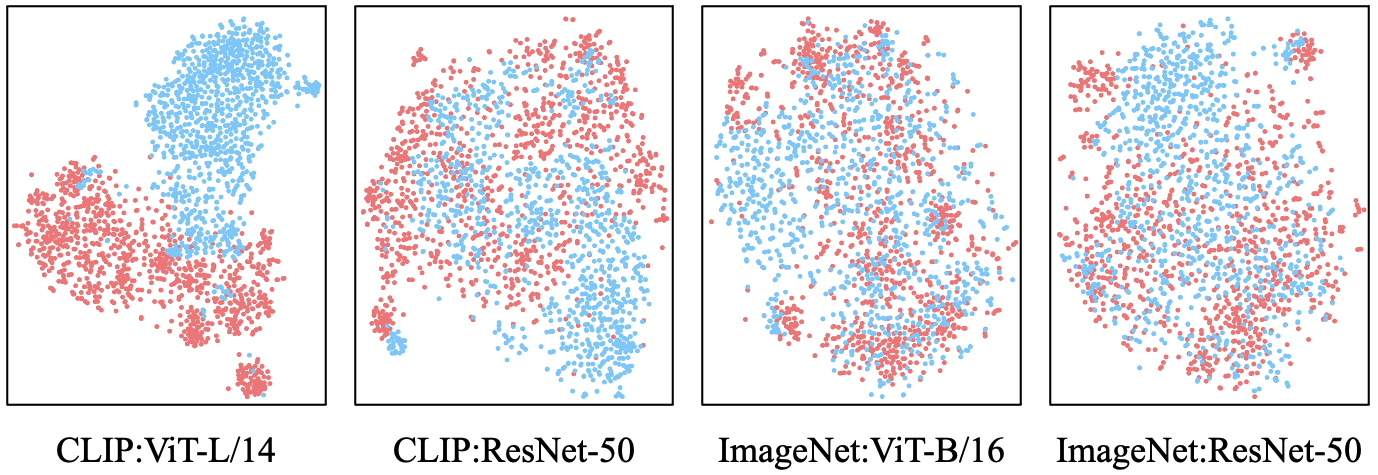}
    \caption{t-SNE visualization of real (red) and fake (blue) images using the feature space of different image encoders. CLIP:ViT's feature space best separates the real features from fake.}
    \label{fig:tsne}
    \vspace{-1em}
\end{figure}

Our approach, on the other hand, show a \emph{drastically better generalization performance} in detecting real/fake images. We observe this first by considering models from the same family as the training domain, i.e., GANs, where our NN variants and linear probing achieve an average accuracy of $\sim$93\% and $\sim$95\% respectively, while the best performing baseline, trained deep networks - Blur+JPEG(0.5) achieves $\sim$85\% (improvements of \textbf{+8-10\%}). This discrepancy in performance becomes more pronounced when considering unseen methods such as diffusion (LDM+Guided+Glide) and autoregressive models (DALL-E), where our NN variants and linear probing achieve 82-84\% average accuracy and $\sim$82\% respectively compared to 53-58\% by trained deep networks variants \cite{cnn-detect} (improvements of \textbf{+25-30\%}). In terms of average precision, the best version of the trained deep network's AP is very high when tested on models from the same GAN family, 94.19 mAP, but drops when tested on unseen diffusion and autoregressive models, 75.51 mAP. Our NN variants and linear probing maintain a high AP both within the same (GAN) family domain, 96.36 and 99.31 mAP, and on the diffusion and autoregressive models, 90.58 and 95.00 mAP, resulting in an improvement of about \textbf{+15-20 mAP} for those unseen models.  

Also, the performance of our NN remains similar even if one varies the voting pool size from $k$=1 to $k$=9. This is good, as it shows that our method is not too sensitive to this hyperparameter in nearest neighbor search. Performing linear classification on that same feature space of CLIP:ViT encoder (Ours LC) preserves, and sometimes enhances the generalization ability of nearest neighbor classification.


In sum, these results clearly demonstrate the advantage of our approach of using the feature space of a frozen, pre-trained network that is \emph{blind to the downstream task of real-vs-fake classification}.


\vspace{-2pt}
\subsection{Allowing the trained classifier to cheat} 
\vspace{-2pt}

As described in Sec.~\ref{sec:eval_metrics}, we experiment with an oracle version of the trained classifier baseline \cite{cnn-detect}, where the threshold of the classifier is tuned directly on each \emph{test set}. Even this flexibility, where the network essentially \emph{cheats(!)} by looking at the test set, does not make the trained classifier perform nearly as well as our approach, especially for models from unseen domains; for example, our nearest neighbor $k=9$ variant achieves an average classification accuracy of 84.25\%, which is \textbf{7.99\% higher than that of the oracle baseline} (76.26\%).  This shows that the issue with training neural networks for this task is not just the improper threshold at test time. Instead, the trained network fundamentally cannot do much other than look for a certain set of fake patterns; when those patterns are not available, it does not have the tools to look for features pertaining to the real distribution. And that is precisely where we believe the feature space of a \emph{model not trained on this task} has its advantages; when certain (e.g., GAN's) low-level patterns are not found, there will still be other features that could be useful for classification, which was not learned to be ruled out during the real-vs-fake training process.



\vspace{-2pt}
\subsection{Effect of network backbone}
\vspace{-2pt}
So far, we have seen the surprisingly good generalizability of nearest neighbor / linear probing using CLIP:ViT-L/14's feature space. In this section, we study how important this choice is, and what happens if the backbone architecture or pre-training dataset is changed. We experiment with our linear classification variant, and consider the following $<$dataset/task$>$:$<$architecture$>$  settings: (i) CLIP:ViT-L/14, (ii) CLIP:ResNet-50, (iii) ImageNet:ResNet-50, and (iv) ImageNet:ViT-B/16. For each, we again use ProGAN's real/fake image data as the training data.

Fig.~\ref{fig:pretrain_ablation} shows the accuracy of these variants on the same models. The key takeaway is that both the network architecture as well as the dataset on which it was trained on play a crucial role in determining the effectiveness for fake image detection. Visual encoders pre-trained as part of the CLIP system fare better compared to those pre-trained on ImageNet. This could be because CLIP's visual encoder gets to see much more diversity of images, thereby exposing it to a much bigger \emph{real distribution} than a model trained on ImageNet. Beyond that, CLIP is trained to align an image with a caption whereas an ImageNet classification model is trained to align an image with a label. Since a caption naturally presents more information about the image, the features extracted by CLIP need to be more descriptive, as opposed to ImageNet model's features which can focus only on the main object. 
Within CLIP, ViT-L/14 performs better than ResNet-50, which could partly be attributed to its bigger architecture and global receptive field of the attention layers.

We also provide a visual analysis of the pre-trained distributions.  Using each of the four model's feature banks consisting of the same real and fake images from ProGAN, we plot four t-SNE figures and color code the resulting 2-D points using binary (real/fake) labels in Fig.~\ref{fig:tsne}.  CLIP:ViT-L/14's space best separates the real (red) and fake (blue) features, followed by CLIP:ResNet-50. ImageNet:ResNet-50 and ImageNet:ViT-B/16 do not seem to have any proper structure in separating the two classes, suggesting that the pre-training data matters more than the architecture. 


\begin{figure*}[t]
    \centering
    \includegraphics[width=.98\textwidth]{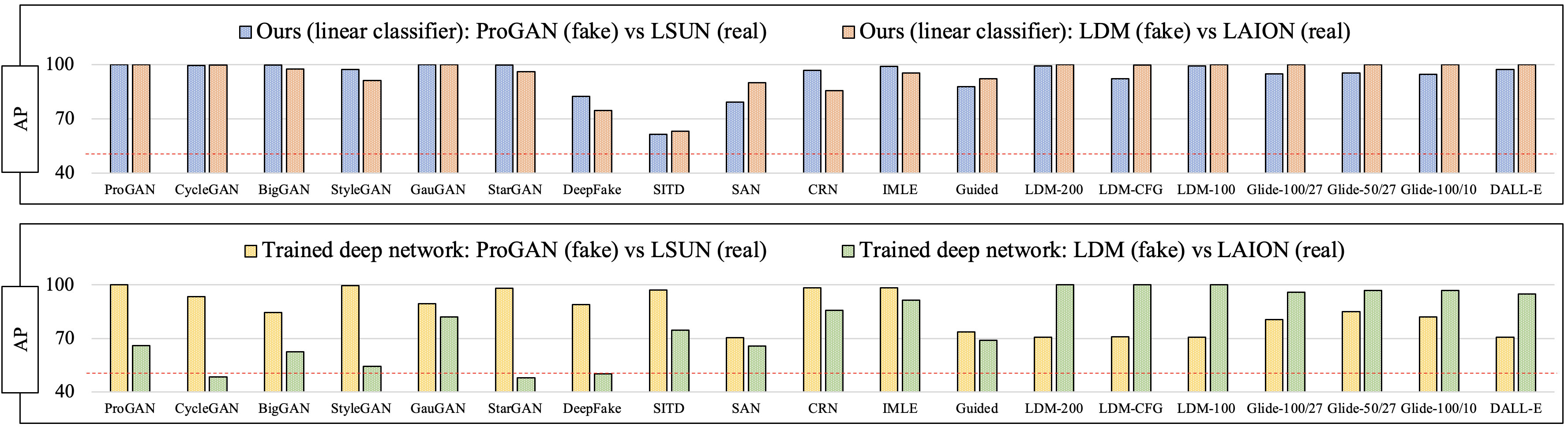}
    \caption{\textbf{Average precision of methods with respect to training data.} Both our linear classifier on CLIP:ViT's features (top) and the baseline trained deep network~\cite{cnn-detect} (bottom) are given access to two different types of training data: (i) $\mathcal{R} =$ LSUN~\cite{yu15lsun} and $\mathcal{F} =$ ProGAN~\cite{progan}, (ii) $\mathcal{R} =$  LAION~\cite{laion} and $\mathcal{F} =$ LDM~\cite{ldm}. Irrespective of the training data source, our linear classifier preserves its ability to generalize well on images from other unseen generative model families. The baseline trained deep network's ability, on the other hand, suffers similarly in both settings.}
    \label{fig:dataset_ablation}
\end{figure*}

\begin{figure*}[t]
    \centering
    \includegraphics[width=.98\textwidth]{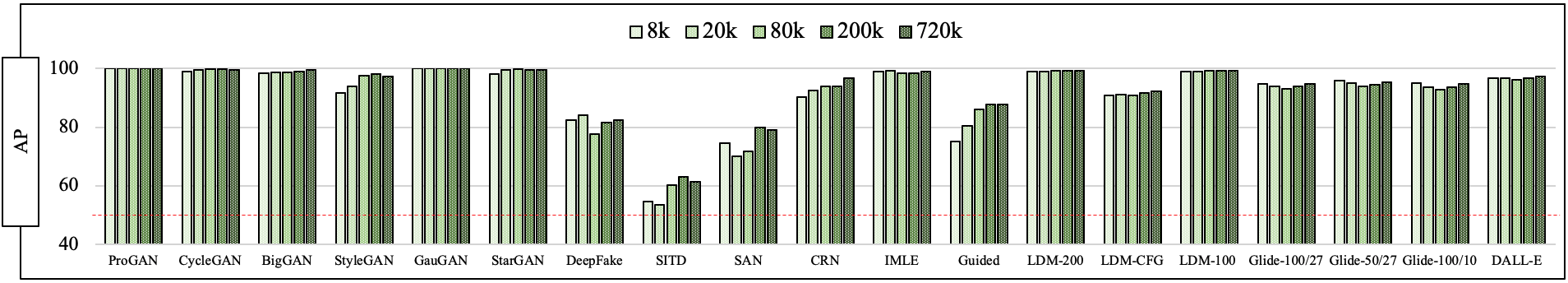}
    \caption{Performance of our linear classifier when the number of available images ($N$) in both $\mathcal{R}$ and $\mathcal{F}$ are varied. Its performance remains more or less similar even after drastically reducing the size of the training set.}
    \label{fig:dataset_size_ablation}
\end{figure*}

\vspace{-1pt}
\subsection{Effect of training data source}\label{sec:data_source}
\vspace{-1pt}

So far, we have used ProGAN as the source of training data.  We next repeat the evaluation setup in Table~\ref{tab:ap} using a pre-trained LDM~\cite{ldm} as the source instead. The real class consists of images from LAION dataset \cite{laion}. Fake images are generated using an LDM 200-step variant using text prompts from the corresponding real images. In total, the dataset consists of 400k real and 400k fake images. 

Fig.~\ref{fig:dataset_ablation} (top) compares our resulting linear classifier to the one created using ProGAN's dataset. Similar to what we have seen so far, access to only LDM's dataset also enables the model to achieve good generalizability. For example, our model can detect images from GAN's domain, which now act as the unseen image generation method, with an average of 97.32 mAP.  In contrast, the trained deep network (Fig.~\ref{fig:dataset_ablation} bottom) performs well only when the target model is from the same generative model family, and fails to generalize in detecting images from GAN variants, 60.17 mAP; i.e., the improvement made by our method for the unseen GAN domain is \textbf{+37.16 mAP}. In summary, with our linear classifier, one can start with ProGAN's data and detect LDM's fake images, or vice versa. This is encouraging from the point of view of image forensics because it tells us that, \emph{so far}, with all the advancements in generative models, there is still a hidden link which connects various fake images.



\vspace{-2pt}
\subsection{Effect of training data size}
\vspace{-2pt}

How much training data does one need for these encouraging results on CLIP:ViT's feature space to hold? So far, our dataset sizes have been 720k/800k for ProGAN/LDM's domains. We use ProGAN's data and experiment with the following overall (real + fake) dataset size: \{8k, 20k, 80k, 200k, 720k\}. As one would expect, performance generally increases with bigger training data; see Fig.~\ref{fig:dataset_size_ablation}. Since the fake class in the training data comes from a GAN, the effect of data size is not felt as much in the GAN's model family (e.g., GauGAN) as it is in the diffusion model family. Still, it is worth noting that even for those unseen generative models, one can reduce the dataset size requirement by $\times$3-4 without a large loss in generalization ability.  

\vspace{-2pt}
\subsection{Visualizing distances in $\phi$'s space}
\vspace{-2pt}

\begin{figure}[]
    \centering
    \includegraphics[width=0.48\textwidth]{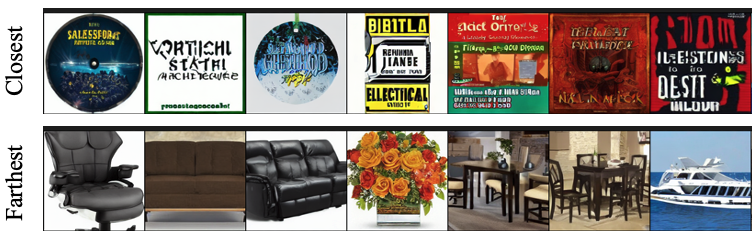}
    \caption{Top/bottom rows indicate fake images from LDM which are closest/farthest from ProGAN's feature space respectively. The images in the top row visually appear to be more fake than images in the bottom row.}
    \label{fig:fake_quality}
    \vspace{-1em}
\end{figure}

We next see whether distances in CLIP:ViT's feature space can tell us something about the visual quality of fake images. We use the same feature bank of ProGAN's fake images and visualize the closest/farthest nearest neighbor fake images from LDM. Fig.~\ref{fig:fake_quality} shows that LDM generated images which are the closest nearest neighbors to ProGAN fakes do tend to be less realistic compared to LDM images which are the farthest nearest neighbors. Overall, this further adds to the utility of the feature space of a large scale model \emph{not trained} for the task of interest.

\begin{figure*}[]
    \centering
    \includegraphics[width=0.98\textwidth]{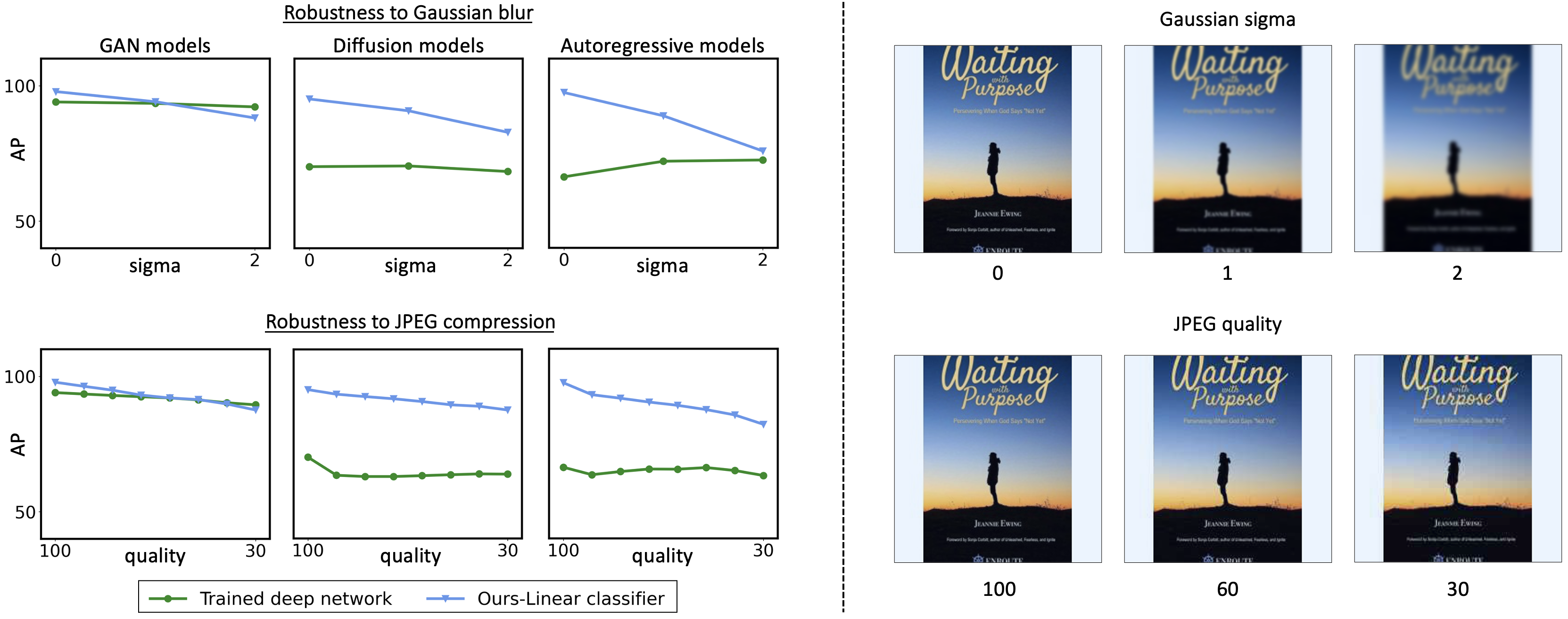}
    \caption{\textbf{Left:} Robustness to different image processing operations. Gaussian blur (top) and JPEG compression (bottom) when tested on fake images of different breeds of generative models (columns). Both our linear classifier and the trained deep network baseline~\cite{cnn-detect} are generally robust to these artifacts, while our performance is significantly higher in absolute AP on unseen generative models (Diffusion and Autoregressive). \textbf{Right:} Effect of Gaussian blur (top) and JPEG compression (bottom), where the corruption strength is increasing from left to right.}
    \label{fig:robustness}
    \vspace{-1em}
\end{figure*}

\subsection{Robustness to post-processing operations} Finally, in order to evade a fake detection system, an attacker might apply certain low-level post-processing operations to their fake images. Therefore, following prior work~\cite{natarajan, Yu_2019_ICCV,zhang2019gan, cnn-detect}, we evaluate how robust our classifiers are to such operations. We study the effects of JPEG compression and Gaussian blurring, and compare our linear classification approach using CLIP:ViT's features (Ours LC) to the trained deep network baseline \cite{cnn-detect}. Fig.~\ref{fig:robustness} (left) shows the results on three types of generative models: GANs (averaged over CycleGAN, BigGAN etc.), diffusion models (averaged over LDM, Glide etc.) and autoregressive model (DALL-E). Note that both our linear classifier as well as the baseline train with jpeg+blur data augmentation on ProGAN real and fakes.  

First, we see that both our method and the baseline are generally robust to blur and jpeg artifacts.  The trained deep network baseline is more consistent in its performance across the varying degrees of blur/compression, but its absolute AP is still much lower than ours when testing on unseen diffusion and autoregressive model fakes.  This makes sense since the baseline trains the whole network (i.e., including the features) to be specifically robust to blur+jpeg effects, whereas our linear classifier only trains the last classification layer while using a pretrained backbone that was not explicitly trained to be robust to blur+jpeg artifacts. More importantly, when visualizing the effect of these operations on images in Fig.~\ref{fig:robustness} (right), we notice that at a certain point (e.g., Gaussian blur sigma=2), the image becomes quite blurry; thus, it's not clear that one would want to degrade image quality to such an extent solely to evade a fake detection system. Therefore, we believe that the need for robustness in such extreme cases becomes less important.

\vspace{-3pt}
\section{Conclusion and Discussion}
\vspace{-2pt}


We studied the problem associated with training neural networks to detect fake images. The analysis paved the way for our simple fix to the problem: using an informative feature space \emph{not trained} for real-vs-fake classification. Performing nearest neighbor / linear probing in this space results in a significantly better generalization ability of detecting fake images, particularly from newer methods like diffusion/autoregressive models. As mentioned in Sec.~\ref{sec:data_source}, these results indicate that even today there is something common between the fake images generated from a GAN and those from a diffusion model. However, what that similarity is remains an open question. And while having a better understanding of that question will be helpful in designing even better fake image detectors, we believe that the generalization benefits of our proposed solutions should warrant them as strong baselines in this line of work.
{\small
\bibliographystyle{ieee_fullname}
\bibliography{main}
}

\clearpage
\appendix

\section*{Appendix}

This document provides additional information complementing the main paper. First, we give more details about the frequency based classification baseline already introduced in the main paper, in Sec.~\ref{sec:zhang}. Next, we discuss the training details of all the baselines presented in this work in Sec.~\ref{sec:train_details}. Following that, we present some additional ablation studies for nearest neighbor search, in Sec.~\ref{sec:supp_ablation}. We then present the breakdown of the generalization performance of a method into accuracy of detecting real and fake images separately, in Tables \ref{tab:real_acc} and \ref{tab:fake_acc}. Finally, we show the precision-recall curves, as was discussed in Table~\ref{tab:table_baseline} in main paper, for two methods: trained deep network~\cite{cnn-detect} and our proposed nearest neighbor search, in Fig.~\ref{fig:pr_curves}.

\section{Details about the frequency spectrum classification}\label{sec:zhang}
In Sec.~\ref{sec:baselines}, we briefly introduce a baseline for classifying real from fake images using their frequency spectrum \cite{zhang2019gan}. In this section, we discuss this baseline in more details. The authors point out that most of the GAN architectures have upsampling layers which introduce checkerboard artifacts in the generated images. These artifacts, authors argue, can be better represented in the frequency space. They consider CycleGAN as their training/evaluation environment, where they train a model on one domain (e.g. real/fake horses) and evaluate on the rest (e.g. CycleGAN generated apples, winter scenes). We follow the same training steps as the authors, but instead train a network on \emph{all} the domains of CycleGAN. Specifically, for each image, either real or fake, we first convert it into its frequency space using 2-D Fourier transform. This frequency image is then fed into a real-vs-fake classification network. The authors use ResNet-34 pretrained on ImageNet in their work, and we follow that recommendation. The network is trained using a two-way cross-entropy loss. During test time, an image is first mapped to its frequency space in a similar manner as the training step, and then fed into the trained network to obtain the real or fake classification decision. 

\begin{figure*}[t!]
    \centering
    \includegraphics[width=1\textwidth]{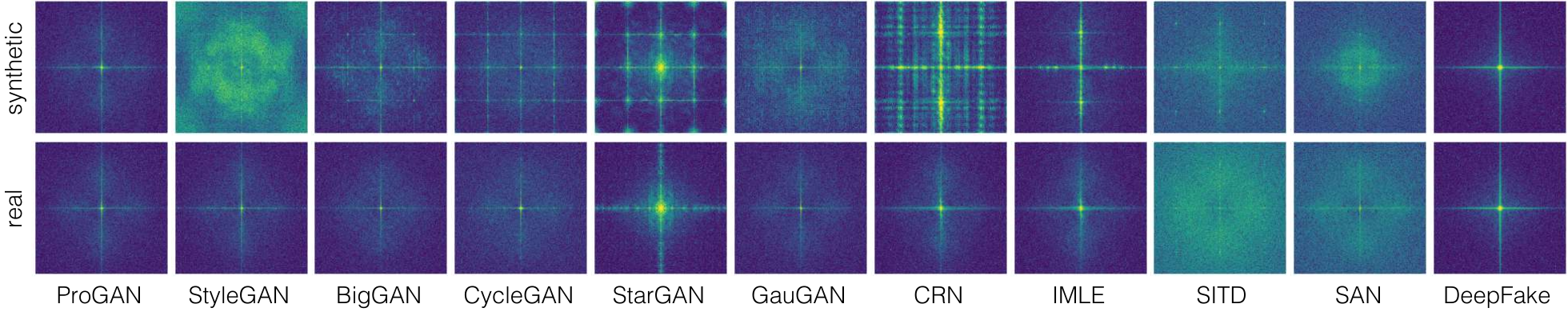}
    \vspace{-15pt}
    \caption{Figure from \cite{cnn-detect}, which shows the frequency spectrum of real and fake images from different generative models.}
    \label{fig:fft_gans}
    \vspace{-5pt}
\end{figure*}

The results are shown in Table~\ref{tab:main_comparison_new} (replicated from the main paper), depicting the average accuracy of classifying real and fake images of a generative model. We see that the method can detect the held out real/fake images from CycleGAN perfectly, which was used for training. This ability is preserved even while detecting real/fake images from StarGAN. For images from all the other generative models, the classification accuracy drops to almost \emph{chance} performance! We think this behavior, where the network works almost perfectly for CycleGAN and StarGAN, and not for others, can be explained by observing the frequency patterns of fake images from different models. This was studied in detail in Fig. 7 of \cite{cnn-detect}, which we are reusing in Fig.~\ref{fig:fft_gans}. We observe that the frequency patterns of fake images from CycleGAN and StarGAN are very similar, with a similar 3x3 grid structure in the middle. And although most of the other generative models also have their own patterns in that space, it is different from that of CycleGAN/StarGAN. So, we believe a similar thing happens here as was shown in Fig. 2 of the main paper. Whenever the classifier finds these 3x3 block patterns, the image is classified as fake; everything else gets classified as real (studied in more detail in Sec.~\ref{sec:rf_acc}). Our method, on the other hand, was never trained to explicitly look for patterns which distinguish one set of fake images from the real ones. Hence, its performance is much more consistent and accurate across different generative models.     

\begin{table*}[t!]
  {\small
    \centering
    \tabcolsep=0.1cm
    \resizebox{1.\linewidth}{!}{
    \begin{tabular}{cc ccccccc cc cc c ccc ccc c c}
    \toprule

        \multirow{2}{*}{\shortstack[c]{Detection\\method}} & \multirow{2}{*}{Variant} & \multicolumn{7}{c}{Generative Adversarial Networks} & \multicolumn{2}{c}{Low level vision} & \multicolumn{2}{c}{Perceptual loss} &\multirow{2}{*}{Guided} & \multicolumn{3}{c}{LDM} & \multicolumn{3}{c}{Glide} & \multirow{2}{*}{DALL-E} & Total \\
    \cmidrule(lr){3-9} \cmidrule(lr){10-11} \cmidrule(lr){12-13} \cmidrule(lr){15-17} \cmidrule(lr){18-20} \cmidrule(lr){22-22}

    & & \shortstack[c]{Pro-\\GAN} & \shortstack[c]{Cycle-\\GAN} & \shortstack[c]{Big-\\GAN} & \shortstack[c]{Style-\\GAN} & \shortstack[c]{Gau-\\GAN} &  \shortstack[c]{Star-\\GAN}   & \shortstack[c]{Deep-\\fakes} & SITD & SAN & CRN & IMLE & & \shortstack[c]{200\\steps} & \shortstack[c]{200\\w/ CFG} & \shortstack[c]{100\\steps} & \shortstack[c]{100\\27} & \shortstack[c]{50\\27} & \shortstack[c]{100\\10} & & \shortstack[c]{Avg.\\acc}
    
    \\ 

    \midrule

\multirow{3}{*}{\shortstack[c]{Trained\\classifier \cite{cnn-detect}}} & Blur+JPEG (0.1) & 99.99 & 85.20 & 70.20 & 85.7 & 78.95 & 91.7 & 53.47 & 66.67 & 48.69 & 86.31 & 86.26 & 60.07 & 54.03 & 54.96 & 54.14 & 60.78 & 63.8 & 65.66 & 55.58 & 69.58 \\
& Blur+JPEG (0.5) & 100.0 & 80.77 & 58.98 & 69.24 & 79.25 & 80.94 & 51.06 & 56.94 & 47.73 & 87.58 & 94.07 & 51.90 & 51.33 & 51.93 & 51.28 & 54.43 & 55.97 & 54.36 & 52.26 & 64.73 \\
& Oracle$^{*}$ (B+J 0.5) & 100.0 & 90.88 & 82.40 & 93.11 & 93.52 & 87.27 & 62.48 & 76.67 & 57.04 & 95.28 & 96.93 & 65.20 & 63.15 & 62.39 & 61.50 & 65.36 & 69.52 & 66.18 & 60.10 & 76.26 \\ 
& ViT:CLIP (B+J 0.5) & 98.94 & 78.80 & 60.62 & 60.56 & 66.82 & 62.31 & 52.28 & 65.28 & 47.97 & 64.09 & 79.54 & 50.66 & 50.74 & 51.04 & 50.76 & 52.15 & 53.07 & 52.06 & 53.18 & 60.57 \\ 

\midrule

\multirow{2}{*}{\shortstack[c]{Patch\\classifier \cite{patchforensics}}} & ResNet50-Layer1  &  94.38 & 67.38 & 64.62 & 82.26 & 57.19 & 80.29 & 55.32 & 64.59 & 51.24 & 54.29 & 55.11 & 65.14 & 79.09 & 76.17 & 79.36 & 67.06 & 68.55 & 68.04 & 69.44 & 68.39\\
& Xception-Block2 &  75.03 & 68.97 & 68.47 & 79.16 & 64.23 & 63.94 & 75.54 & 75.14 & 75.28 & 72.33 & 55.3 & 67.41 & 76.5 & 76.1 & 75.77 & 74.81 & 73.28 & 68.52 & 67.91 & 71.24\\

\midrule

\shortstack[c]{Freq-spec \cite{zhang2019gan}} & CycleGAN & 49.90 & 99.90 & 50.50 & 49.90 & 50.30 & 99.70 & 50.10 & 50.00 & 48.00 & 50.60 & 50.10 & 50.00 & 50.40 & 50.40 & 50.30 & 51.70 & 51.40 & 50.40 & 50.00 & 55.45\\

\midrule

\multirow{3}{*}{\shortstack[c]{Nearest\\neighbor\\(\textbf{Ours})}} & $k=1$ & 99.58 & 94.70 & 86.95 & 80.24 & 96.67 & 98.84 & 80.9 & 71.0 & 56.0 & 66.3 & 76.5 & 68.76 & 89.56 & 68.99 & 89.51 & 86.44 & 88.02 & 87.27 & 77.52 & 82.30 \\
& $k=3$ & 99.58 & 95.04 & 87.63 & 80.55 & 96.94 & 98.77 & 83.05 & 71.5 & 59.5 & 66.69 & 76.87 & 70.02 & 90.37 & 70.17 & 90.57 & 87.84 & 89.34 & 88.78 & 79.29 & 83.28 \\
& $k=5$ & 99.60 & 94.32 & 88.23 & 80.60 & 97.00 & 98.90 & 83.85 & 71.5 & 60.0 & 67.04 & 78.02 & 70.55 & 90.89 & 70.97 & 91.01 & 88.42 & 90.07 & 89.60 & 80.19 & 83.72  \\
& $k=9$ & 99.54 & 93.49 & 88.63 & 80.75 & 97.11 & 98.97 & 84.5 & 71.5 & 61.0 & 69.27 & 79.21 & 71.06 & 91.29 & 72.02 & 91.29 & 89.05 & 90.67 & 90.08 & 81.47 & \textbf{84.25} \\

    \bottomrule
    \end{tabular}}
    }
    \vspace{-1.5mm}
    \caption{\textbf{Generalization results.} Classification accuracy of different methods for detecting real/fake images. Numbers indicate the average accuracy over real and fake images from a test model. Oracle with $^{*}$ indicates that the method uses the test set to calibrate the confidence threshold. Models outside the GANs column can be considered as the generalizing domain.}
    \label{tab:main_comparison_new}

\end{table*}

\begin{table*}[t!]
  {\small
    \centering
    \tabcolsep=0.1cm
    \resizebox{1.\linewidth}{!}{
    \begin{tabular}{cc ccccccc cc cc c ccc ccc c c}
    \toprule

        \multirow{2}{*}{\shortstack[c]{Detection\\method}} & \multirow{2}{*}{Variant} & \multicolumn{7}{c}{Generative Adversarial Networks} & \multicolumn{2}{c}{Low level vision} & \multicolumn{2}{c}{Perceptual loss} &\multirow{2}{*}{Guided} & \multicolumn{3}{c}{LDM} & \multicolumn{3}{c}{Glide} & \multirow{2}{*}{DALL-E} & Total \\
    \cmidrule(lr){3-9} \cmidrule(lr){10-11} \cmidrule(lr){12-13} \cmidrule(lr){15-17} \cmidrule(lr){18-20} \cmidrule(lr){22-22}

    & & \shortstack[c]{Pro-\\GAN} & \shortstack[c]{Cycle-\\GAN} & \shortstack[c]{Big-\\GAN} & \shortstack[c]{Style-\\GAN} & \shortstack[c]{Gau-\\GAN} &  \shortstack[c]{Star-\\GAN}   & \shortstack[c]{Deep-\\fakes} & SITD & SAN & CRN & IMLE & & \shortstack[c]{200\\steps} & \shortstack[c]{200\\w/ CFG} & \shortstack[c]{100\\steps} & \shortstack[c]{100\\27} & \shortstack[c]{50\\27} & \shortstack[c]{100\\10} & & \shortstack[c]{Avg.\\acc}
    
    \\ 

    \midrule

\multirow{3}{*}{\shortstack[c]{Trained\\classifier \cite{cnn-detect}}} & Blur+JPEG (0.1) & 100.0 & 91.52 & 93.5 & 99.92 & 93.02 & 96.75 & 99.96 & 86.67 & 98.5 & 72.75 & 72.75 & 93.77 & 99.14 & 99.14 & 99.14 & 99.14 & 99.14 & 99.14 & 99.14 & 94.37 \\
& Blur+JPEG (0.5) & 100.0 & 98.64 & 99.05 & 99.95 & 99.40 & 99.30 & 99.45 & 100.0 & 100.0 & 99.22 & 99.22 & 99.14 & 99.61 & 99.61 & 99.61 & 99.61 & 99.61 & 99.61 & 99.61 & 99.50 \\
& Oracle$^{*}$ (B+J 0.5) & 100.0 & 91.98 & 77.15 & 93.22 & 91.76 & 89.99 & 76.51 & 85.0 & 53.5 & 94.45 & 97.12 & 61.43 & 51.71 & 54.06 & 54.06 & 67.98 & 68.23 & 64.47 & 60.83 & 75.44 \\ 
& ViT:CLIP (B+J 0.5) & 99.90 & 96.14 & 98.10 & 99.65 & 99.22 & 99.10 & 99.11 & 100.00 & 100.00 & 98.34 & 98.34 & 98.33 & 97.57 & 97.57 & 97.57 & 97.57 & 97.57 & 97.57 & 97.57 & 98.37 \\ 

\midrule

\multirow{2}{*}{\shortstack[c]{Patch\\classifier \cite{patchforensics}}} & ResNet50-Layer1  &  95.30 & 65.56 & 61.35 & 85.95 & 49.88 & 75.83 & 89.21 & 43.48 & 47.24 & 12.25 & 12.25 & 61.34 & 84.86 & 84.86 & 84.86 & 84.86 & 84.86 & 84.86 & 84.86 & 68.08 \\
& Xception-Block2 &  98.34 & 60.62 & 59.36 & 90.87 & 45.71 & 86.63 & 89.89 & 26.97 & 48.05 & 11.56 & 11.56 & 59.12 & 79.31 & 79.31 & 79.31 & 79.31 & 79.31 & 79.31 & 79.31 & 65.46\\

\midrule

\shortstack[c]{Freq-spec \cite{zhang2019gan}} & CycleGAN & 99.80 & 99.80 & 99.10 &  99.90 & 99.80 & 99.30 & 100.0 & 100.0 & 100.0 & 99.80 & 99.80 & 99.60 & 99.40 & 99.40 & 99.50 & 99.40 & 99.50 & 99.40 & 99.60 & 99.60\\

\midrule

\multirow{3}{*}{\shortstack[c]{Nearest\\neighbor\\(\textbf{Ours})}} & $k=1$ & 99.15 & 90.46 & 94.8 & 99.55 & 96.5 & 99.4 & 96.4 & 76.0 & 95.0 & 97.43 & 97.43 & 94.55 & 93.43 & 93.43 & 93.43 & 93.43 & 93.43 & 93.43 & 93.43  & 94.24 \\
& $k=3$ & 99.15 & 90.39 & 93.85 & 99.67 & 96.5 & 99.5 & 96.7 & 73.0 & 96.0 & 97.05 & 97.05 & 94.6 & 93.03 & 93.03 & 93.03 & 93.03 & 93.03 & 93.03 & 93.03 & 93.92  \\
& $k=5$ & 99.2 & 88.95 & 93.3 & 99.63 & 96.18 & 99.35 & 96.0 & 67.0 & 96.0 & 96.57 & 96.57 & 94.22 & 92.83 & 92.83 & 92.83 & 92.83 & 92.83 & 92.83 & 92.83 & 93.30  \\
& $k=9$ & 99.08 & 87.21 & 92.55 & 99.63 & 95.88 & 99.35 & 96.0 & 61.0 & 95.0 & 96.47 & 96.47 & 93.34 & 92.39 & 92.39 & 92.39 & 92.39 & 92.39 & 92.39 & 92.39 & 92.56 \\

    \bottomrule
    \end{tabular}}
    }
    \vspace{-1.5mm}
    \caption{\textbf{Accuracy of detecting real images.} For each generative model (column), we consider its corresponding real images and test how frequently a classifier (row) correctly predicts it as real.}
    \label{tab:real_acc}

\end{table*}

\begin{table*}[t!]
  {\small
    \centering
    \tabcolsep=0.1cm
    \resizebox{1.\linewidth}{!}{
    \begin{tabular}{cc ccccccc cc cc c ccc ccc c c}
    \toprule

        \multirow{2}{*}{\shortstack[c]{Detection\\method}} & \multirow{2}{*}{Variant} & \multicolumn{7}{c}{Generative Adversarial Networks} & \multicolumn{2}{c}{Low level vision} & \multicolumn{2}{c}{Perceptual loss} &\multirow{2}{*}{Guided} & \multicolumn{3}{c}{LDM} & \multicolumn{3}{c}{Glide} & \multirow{2}{*}{DALL-E} & Total \\
    \cmidrule(lr){3-9} \cmidrule(lr){10-11} \cmidrule(lr){12-13} \cmidrule(lr){15-17} \cmidrule(lr){18-20} \cmidrule(lr){22-22}

    & & \shortstack[c]{Pro-\\GAN} & \shortstack[c]{Cycle-\\GAN} & \shortstack[c]{Big-\\GAN} & \shortstack[c]{Style-\\GAN} & \shortstack[c]{Gau-\\GAN} &  \shortstack[c]{Star-\\GAN}   & \shortstack[c]{Deep-\\fakes} & SITD & SAN & CRN & IMLE & & \shortstack[c]{200\\steps} & \shortstack[c]{200\\w/ CFG} & \shortstack[c]{100\\steps} & \shortstack[c]{100\\27} & \shortstack[c]{50\\27} & \shortstack[c]{100\\10} & & \shortstack[c]{Avg.\\acc}
    
    \\ 

    \midrule

\multirow{3}{*}{\shortstack[c]{Trained\\classifier \cite{cnn-detect}}} & Blur+JPEG (0.1) & 99.98 & 78.88 & 46.9 & 71.49 & 64.88 & 86.64 & 6.82 & 46.67 & 3.2 & 99.87 & 99.76 & 26.36 & 8.92 & 10.77 & 9.15 & 22.43 & 28.47 & 24.24 & 12.02 & 44.60 \\
& Blur+JPEG (0.5) & 100.0 & 62.91 & 18.9 & 38.52 & 59.1 & 62.58 & 2.52 & 13.89 & 0.0 & 75.95 & 88.92 & 4.67 & 3.05 & 4.26 & 2.96 & 9.25 & 12.34 & 9.1 & 4.9 & 30.20 \\
& Oracle$^{*}$ (B+J 0.5) & 100.0 & 89.78 & 87.65 & 92.99 & 95.28 & 84.54 & 48.41 & 68.33 & 60.27 & 96.1 & 96.74 & 68.96 & 74.6 & 70.72 & 70.95 & 62.75 & 70.8 & 67.89 & 59.37 & 77.16 \\ 
& ViT:CLIP (B+J 0.5) & 97.98 & 61.47 & 23.15 & 21.47 & 34.42 & 25.51 & 5.3 & 30.56 & 0.46 & 29.85 & 60.75 & 2.99 & 3.91 & 4.51 & 3.94 & 6.73 & 8.57 & 6.55 & 8.78 & 22.99\\ 

\midrule

\multirow{2}{*}{\shortstack[c]{Patch\\classifier \cite{patchforensics}}} & ResNet50-Layer1  &  93.45 & 69.20 & 67.90 & 78.56 & 64.51 & 84.74 & 21.31 & 85.70 & 54.90 & 96.33 & 97.96 & 68.94 & 73.32 & 67.48 & 73.86 & 49.26 & 52.23 & 51.22 & 54.02 & 68.68\\
& Xception-Block2 &  96.66 & 78.82 & 77.58 & 84.45 & 76.45 & 96.42 & 42.57 & 94.03 & 54.48 & 98.73 & 99.04 & 72.31 & 73.94 & 71.29 & 74.56 & 51.54 & 56.44 & 52.00 & 58.99 & 74.22\\

\midrule

\shortstack[c]{Freq-spec \cite{zhang2019gan}} & CycleGAN & 0.20 & 100.0 & 1.80 & 0.0 & 0.90 & 100.0 & 0.0 & 0.0 & 0.40 & 1.30 & 0.50 & 0.40 & 1.30 & 1.40 & 1.10 & 3.90 & 3.30 & 1.30 & 0.50 & 11.50\\

\midrule

\multirow{3}{*}{\shortstack[c]{Nearest\\neighbor\\(\textbf{Ours})}} & $k=1$ & 100.0 & 98.94 & 79.1 & 60.92 & 96.84 & 97.4 & 65.4 & 66.0 & 17.0 & 35.16 & 55.56 & 42.97 & 85.69 & 44.55 & 85.59 & 79.45 & 82.6 & 81.12 & 61.6 & 70.30 \\
& $k=3$ & 100.0 & 99.7 & 81.4 & 61.43 & 97.38 & 98.05 & 69.4 & 70.0 & 23.0 & 36.32 & 56.69 & 45.44 & 87.71 & 47.31 & 88.11 & 82.65 & 85.65 & 84.54 & 65.55 & 72.64 \\
& $k=5$ & 100.0 & 99.7 & 83.15 & 61.58 & 97.82 & 98.45 & 71.7 & 76.0 & 24.0 & 37.51 & 59.48 & 46.88 & 88.96 & 49.12 & 89.19 & 84.01 & 87.31 & 86.36 & 67.56 & 74.14  \\
& $k=9$ & 100.0 & 99.77 & 84.7 & 61.88 & 98.34 & 98.6 & 73.0 & 82.0 & 27.0 & 42.06 & 61.94 & 48.77 & 90.2 & 51.65 & 90.2 & 85.7 & 88.94 & 87.77
 & 70.55 &  75.95 \\

    \bottomrule
    \end{tabular}}
    }
    \vspace{-1.5mm}
    \caption{\textbf{Accuracy of detecting fake images.} For each generative model (column), we consider its corresponding fake images and test how frequently a classifier (row) correctly predicts it as fake.}
    \label{tab:fake_acc}

\end{table*}

\section{Training details}\label{sec:train_details}
In this section, we provide the details surrounding training of different baselines used in our work. For training the image-level classifier proposed in \cite{cnn-detect}, we use the official code repository given by the authors.\footnote{Code and pre-trained models taken from \href{https://github.com/PeterWang512/CNNDetection}{here}.} The models indicated by Blur+JPEG (0.1) and Blur+JPEG (0.5) are the official pre-trained models released by the authors. We train the ViT:CLIP version of this baseline by using Blur + JPEG data augmentation with 0.5 probability. The network is trained with a batch size of 256 and learning rate of 5e-5. During test time, we do not apply any Blur or JPEG augmentation. For training the patch-based classifier \cite{patchforensics}, we use the network architectures of ResNet50 and Xception given in the official implementation by the authors.\footnote{Code taken from \href{https://github.com/chail/patch-forensics}{here}.} We then train the resulting models in the same way as \cite{cnn-detect} (e.g. same batch size, learning rate). The training objective for each patch is defined using a two-way cross-entropy loss. For both \cite{cnn-detect} and \cite{patchforensics}, we follow the authors' respective recommendation of terminating the training process, which involves tracking the accuracy on a held out validation set. For training the classifier in the frequency space \cite{zhang2019gan}, we use the official code given by the authors.\footnote{Code taken from \href{https://github.com/ColumbiaDVMM/AutoGAN}{here}.} The classifier is trained on all the domains of CycleGAN considered by the authors (14 total). During both training and testing, an image is first converted into its frequency space using 2-D Fourier transform. We follow the training hyperparametrs given in the official implementation. For training the classifier on co-occurence matrices of real/fake images \cite{natarajan}, we contacted the authors and followed their guidelines. The network architecture is kept the same as prescribed in the paper, which is trained without using any data augmentations. Lastly, when training our proposed linear classifier, we make use of blur+jpeg data augmentations, as suggested by \cite{cnn-detect}; i.e., any real/fake image is first augmented before being passed to the CLIP:ViT encoder ($\phi$). We use the same stopping criteria for this method as done for \cite{cnn-detect, patchforensics}. 

\paragraph{Tuning the threshold:} As discussed in Section 5.3 of the main paper, the image-level classifier from \cite{cnn-detect} makes a real/fake decision based on the threshold that we choose. For all non-oracle baselines, we consider the validation set of ProGAN to decide the threshold. We pass all the real and fake images in that set through the trained network and obtain the corresponding scores after sigmoid operation, which serve as candidate thresholds. We iterate through each of them and find the threshold which results in highest accuracy on that same validation set. Typically, we find that any value between [0+$\epsilon$, 1-$\delta$] (where $\epsilon$ and $\delta$ are very small values) results in the same almost perfect accuracy. So, we simply use the middle of those two extremes, $\sim$0.5 as our threshold. For oracle version, we repeat the same process as above, but find those candidate thresholds using the test set itself. In other words, the ideal threshold for each test set is different for each generative model.

\section{Additional ablation for NN}\label{sec:supp_ablation}
In this section, we continue our study of the different components which affect the generalization ability of nearest neighbors.
\subsection{Effect of different layers of CLIP as $\phi$}
In all the experiments in the main paper, we have used the last layer of CLIP as the feature space for performing nearest neighbor search. Here, we study the importance of that choice and see what happens if we choose some different layer. In particular, CLIP:ViT has a total of \ut{24} layers, with the 24th being the last layer that we have used so far. We consider the following layers \{$L_0$, $L_8$, $L_{16}$, $L_{24}$\} where $L_{i}$ indicates the $i^{th}$ layer, and perform nearest neighbor classification with $k=1$.

\begin{figure*}[t]
    \centering
    \includegraphics[width=0.98\textwidth]{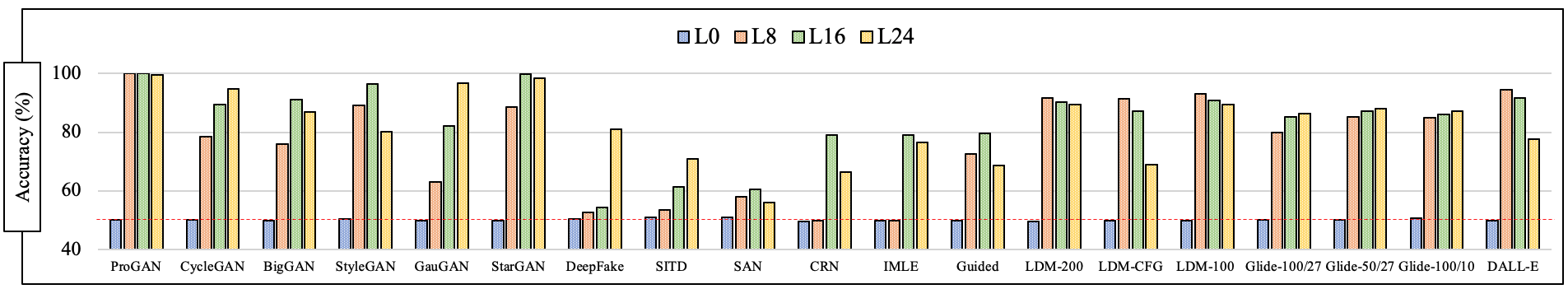}
    \caption{Effect of different layer choices in CLIP:ViT on the generalization ability. $L_0$ corresponds to the first layer in CLIP:ViT and $L_{24}$ corresponds to the last layer which has been used in all of our experiments so far.}
    \label{fig:layer_abl}
    \vspace{-1em}
\end{figure*}

Fig.~\ref{fig:layer_abl} shows the results, where we see that as long as the layer chosen is not the very first one, which is almost equivalent to performing nearest neighbor search in pixel space, the generalization ability remains decently consistent. Similar to how nearest neighbor search is shown to be robust to the size of voting pool (e.g. $k=1$ or $k=9$) in Sec. 6.1 of main paper, Fig.~\ref{fig:layer_abl} adds another dimension of robustness, where one does not need to worry too much about which layer to use as $\phi$.

\subsection{Effect of dataset diversity}
The default dataset that we have used throughout the main paper (e.g. Table 2) comes from 20 different ProGANs trained on different domains, i.e., 20 LSUN object classes of real/fake images. In this section, we study how important this diversity is. We consider four variants of datasets, comprising of 2, 4, 8 and 20 classes. We construct nearest neighbor search separately with each of these as the training source.

\begin{figure*}[t]
    \centering
    \includegraphics[width=0.98\textwidth]{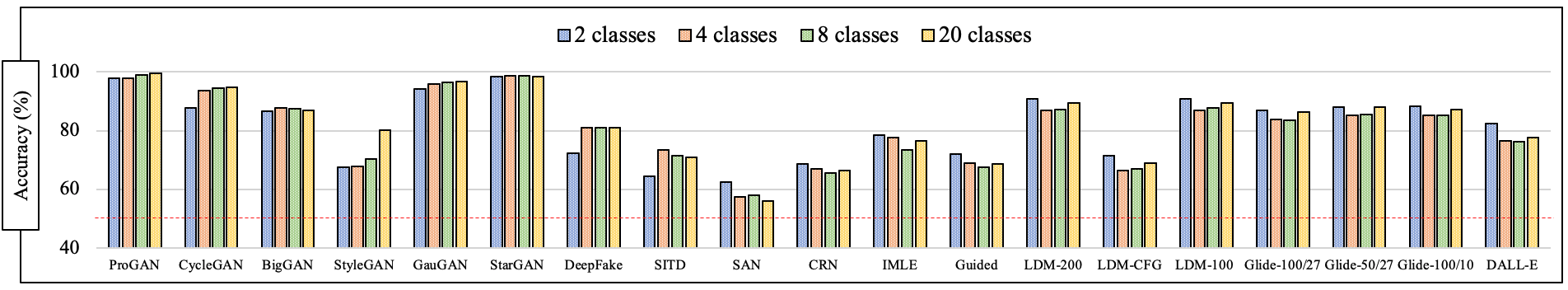}
    \caption{Effect of number of classes in the training set $\mathcal{D}$ on the generalization ability. `2 classes' means that the training set consists of 2 object classes of real and 2 object classes of corresponding fake images.}
    \label{fig:class_abl}
    \vspace{-1em}
\end{figure*}

Fig.~\ref{fig:class_abl} shows the results, where we see that for most of the generative models, all variations, even when the training data consists of real/fake images from just 2 classes, performs decently well. This is in contrast to the analogous analysis done in \cite{cnn-detect}, where the authors found that training the image-level classifier on reduced dataset diversity, especially when there are only 2 classes, hurts generalization ability of the resulting detector.

\subsection{Using data from multiple generative models}
In all the experiments so far, we have restricted the training data to be from a single source (either ProGAN or LDM). In this section, we study what happens if we relax that constraint. In particular, we compare our method (nearest neighbor) with image-level classifier when both methods have access to real and fake data from two different domains. $\mathcal{R}$ = \{$R_{LSUN} \cup R_{LAION}$\}  and $\mathcal{F}$ = \{$F_{ProGAN} \cup F_{LDM}$\}.  

\begin{figure*}[t!]
    \centering
    \includegraphics[width=0.98\textwidth]{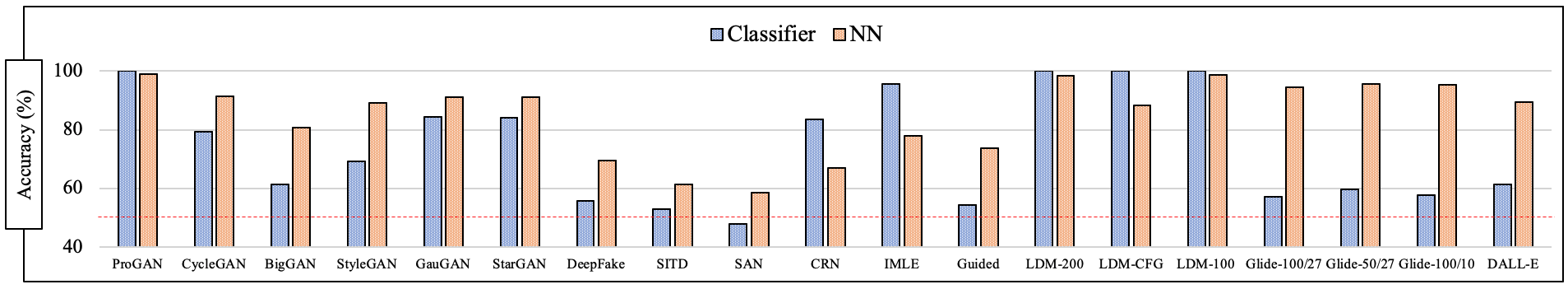}
    \caption{Comparison of a trained classifier \cite{cnn-detect} vs nearest neighbor when both have access to real/fake images from \emph{two} domains: ProGAN and LDM. Note that the trained classifier does not generalize well to domains not seen during training (e.g. Glide models).}
    \label{fig:2x_data}
    \vspace{-1em}
\end{figure*}

\begin{figure*}[h]
    \centering
    \includegraphics[width=0.98\textwidth]{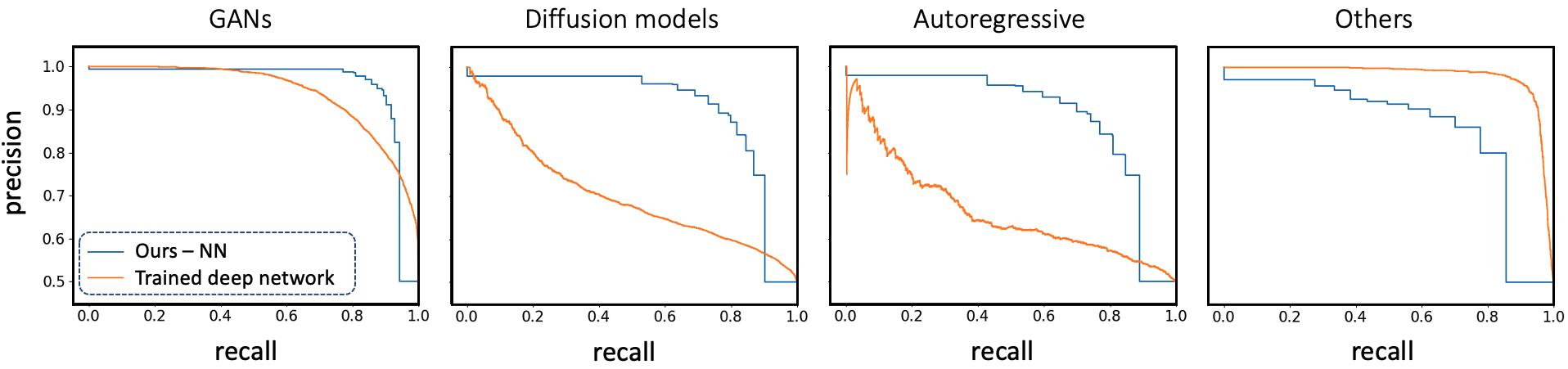}
    \caption{Precision-recall curves for our proposed nearest neighbor classifier ($k$=1) and the baseline trained deep network~\cite{cnn-detect}, for the task of binary real/fake classification. Similar to Fig.~\ref{fig:robustness} in the main paper, the plots are computed by averaging results over specific generative model families; e.g., GANs represent an average over CycleGAN, BigGAN etc. The training data for both the classifiers consist of LSUN (real) and ProGAN's images (fake). We can clearly see the drop in performance (lower area under curve) for the trained deep network baseline when tested on diffusion/autoregressive models.}
    \label{fig:pr_curves}
    \vspace{-1em}
\end{figure*}

Fig.~\ref{fig:2x_data} shows the results, where we see that the story of the baseline classifier does not change: it performs well on the domains seen during training (ProGAN and LDM variants), but its performance remains poor on the other unseen domains (e.g., DALL-E). Nearest neighbor, on the other hand, has a much more consistent behavior across different models. This shows that the difference in generalization abilities of our method compared to a trained classifier cannot simply be solved by giving access to more sources of real/fake images.

\section{Accuracy breakdown of real and fake classes}\label{sec:rf_acc}
Lastly, we break down the performance of different methods, i.e., Table~\ref{tab:main_comparison_new} into performance on real (Table~\ref{tab:real_acc}) and fake images (Table~\ref{tab:fake_acc}) associated with different generative models. This breakdown helps us understand the particular way in which a detection method fails to work. In particular, we see that an image level classifier \cite{cnn-detect} works fine in detecting real/fake images when they are within the GAN domain. When tested on, for example, images from latent diffusion models, the network starts classifying everything as real. Hence, the classification accuracy on real images remains high, but accuracy on fake images drops down drastically. Similarly, the classifier trained on frequency spectrum \cite{zhang2019gan} of images works well on CycleGAN images likely because it can use the presence/absence of the 3x3 grid pattern. StarGAN's fake images preserve that pattern, and hence the classifier has no issues detecting those fake images as such. However, no other generative model seems to generate images which have those 3x3 grid structures. Because of this, the network classifies everything as real. Hence, the classification accuracy on fake images goes down to almost 0\%. In contrast, performing classification using nearest neighbor does not result in such a big discrepancy, where the network keeps similar outputs irrespective of whether an image was real or fake.

\end{document}